\documentclass[arxiv]{article}

\usepackage[utf8]{inputenc} 
\usepackage[T1]{fontenc}    
\usepackage{hyperref}       
\usepackage{url}            
\usepackage{booktabs}       
\usepackage{amsfonts}       
\usepackage{nicefrac}       
\usepackage{microtype}      
\usepackage{xcolor}         

\usepackage{soul}
\usepackage{url}
\usepackage{epstopdf}
\usepackage{float}
\usepackage{wrapfig}
\usepackage{graphicx}
\usepackage{amsmath}
\usepackage{amssymb}
\usepackage{booktabs}
\usepackage{algorithm}
\usepackage[noend] {algorithmic}
\usepackage{enumitem}
\usepackage{booktabs} 
\usepackage{graphicx,array}
\usepackage{color,soul,xspace}

\usepackage{comment}
\usepackage{epsfig}
\usepackage{mathrsfs,mdwlist,enumitem}

\newcommand{\lnn}{\textsc{Lnn}\xspace}
\newcommand{\hnn}{\textsc{Hnn}\xspace}

\newcommand{\MLP}{\texttt{MLP}\xspace}
\newcommand{\sqp}{\texttt{squareplus}\xspace}

\newcommand{\CV}{\mathcal{V}\xspace}
\newcommand{\CE}{\mathcal{E}\xspace}

\newcommand{\cW}{\mathbf{W}\xspace}

\newcommand{\ch}{\mathbf{h}\xspace}

\newcommand{\cz}{\mathbf{z}\xspace}

\newcommand{\gnode}{\textsc{Gnode}\xspace}
\newcommand{\node}{\textsc{Node}\xspace}
\newcommand{\lgn}{\textsc{Lgn}\xspace}
\newcommand{\hgn}{\textsc{Hgn}\xspace}
\newcommand{\mcgnode}{\textsc{MCGnode}\xspace}
\newcommand{\cgnode}{\textsc{CGnode}\xspace}
\newcommand{\cdgnode}{\textsc{CDGnode}\xspace}

\setlist{nolistsep,leftmargin=*}

\newtheorem{thm}{\textbf{Theorem}}

\newcommand{\rev}[1]{\textcolor{black}{#1}}

\usepackage{microtype}
\usepackage{graphicx}
\usepackage{subcaption}
\usepackage{booktabs} 

\usepackage[a4paper, total={6in, 8in}]{geometry}

\title{\textbf{Enhancing the Inductive Biases of Graph Neural ODE for Modeling Dynamical Systems}}

\usepackage{authblk}
\author{Suresh Bishnoi, Ravinder Bhattoo, Jayadeva, Sayan Ranu, N. M. Anoop Krishnan \thanks{SB: School of Interdisciplinary Research, SR: Department of Computer Science, NMAK and RB: Department of Civil Engineering, J: Department of Electrical Engineering, SR and NMAK: Yardi School of Artificial Intelligence (joint appointment).}\\
Indian Institute of Technology Delhi, Hauz Khas, New Delhi, India 110016\\
\texttt{\{srz208500,cez177518,jayadeva,sayanranu,krishnan\}@iitd.ac.in}
}

\begin{document}

\maketitle

\vspace{-0.30in}

\begin{abstract}
Neural networks with physics-based inductive biases such as Lagrangian neural networks (\textsc{Lnn}), and Hamiltonian neural networks (\textsc{Hnn}) learn the dynamics of physical systems by encoding strong inductive biases. Alternatively, Neural ODEs with appropriate inductive biases have also been shown to give similar performances. However, these models, when applied to particle-based systems, are transductive in nature and hence, do not generalize to large system sizes. In this paper, we present a graph-based neural ODE, \gnode, to learn the time evolution of dynamical systems. Further, we carefully analyze the role of different inductive biases on the performance of \gnode. We show that similar to \textsc{Lnn} and \textsc{Hnn}, encoding the constraints explicitly can significantly improve the training efficiency and performance of \gnode{} significantly. Our experiments also assess the value of additional inductive biases, such as Newton’s third law, on the final performance of the model. We demonstrate that inducing these biases can enhance the performance of the model by orders of magnitude in terms of both energy violation and rollout error. Interestingly, we observe that the \gnode{} trained with the most effective inductive biases, namely \mcgnode, outperforms the graph versions of \textsc{Lnn} and \textsc{Hnn}, namely, Lagrangian graph networks (\textsc{Lgn}) and Hamiltonian graph networks (\textsc{Hgn}) in terms of energy violation error by \(\sim\)4 orders of magnitude for a pendulum system, and \(\sim\)2 orders of magnitude for spring systems. These results suggest that NODE-based systems can give competitive performances with energy-conserving neural networks by employing appropriate inductive biases.
\end{abstract}
\vspace{-0.10in}
\section{Introduction and related works}
\vspace{-0.10in}
Learning the dynamics of physical systems is a challenging problem that has relevance in several areas of science and engineering such as astronomy (motion of planetary systems), biology (movement of cells), physics (molecular dynamics), and engineering (mechanics, robotics)~\cite{lavalle2006planning,goldstein2011classical}. The dynamics of a system are typically expressed as a differential equation, solutions of which may require the knowledge of abstract quantities such as force, energy, and drag~\cite{lavalle2006planning,zhong2020dissipative,sanchez2020learning}. From an experimental perspective, the real observable for a physical system is its trajectory represented by the position and velocities of the constituent particles. Thus, learning the abstract quantities required to solve the equation, directly from the trajectory, can extremely simplify the problem of learning the dynamics~\cite{lnn1}.

Infusing physical laws as prior has been shown to improve learning in terms of additional properties such as energy conservation, and symplectic nature~\cite{karniadakis2021physics,lnn2,liu2021physics}. To this extent, three broad approaches have been proposed, namely, \textit{Lagrangian} neural networks (\lnn)~\cite{lnn,lnn1,lnn2}, \textit{Hamiltonian} neural networks (\hnn)~\cite{sanchez2019hamiltonian,greydanus2019hamiltonian,zhong2020dissipative,zhong2021benchmarking}, and \textit{neural ODE} (\node)~\cite{chen2018neural,gruver2021deconstructing}. The learning efficiency of \lnn{}s and \hnn{}s is shown to enhance significantly by employing explicit constraints~\cite{lnn1} \rev{and their inherent structure~\cite{zhong2019symplectic}}. In addition, it has been shown that the superior performance of \hnn{}s and \lnn{}s is mainly due to their second-order bias, and not due to their symplectic or energy conserving bias~\cite{gruver2021deconstructing}. More specifically, an \hnn{} with separable potential ($V(q)$) and kinetic ($T(q,\dot{q})$) energies is equivalent to a second order \node{} of the form $\ddot{q}=\hat{F}(q,\dot{q})$. Thus, a \node with a second-order bias can give similar performances to that of an \hnn~\cite{gruver2021deconstructing}. \rev{Recently, a Bayesian-symbolic approach, which assumes the knowledge of Newtonian mechanics along with symbolic features, has been used to learn the dynamics of physical systems~\cite{xu2021bayesian}.}

An important limitation pervading these models is their \textit{transductive} nature; they need to be trained on each system separately before simulating their dynamics. For instance, a \node trained on $5$-pendulum can work only for $5$-pendulum. This approach is referred to as transductive. In contrast, an inductive model, such as \gnode, when trained on $5$-pendulum ``learns'' the underlying function governing the dynamics at a node and edge level. Hence, a \gnode, once trained, can perform inference on \textit{any} system size that is significantly smaller or larger than the ground truth as we demonstrate later in \S~\ref{sec:experiments}. This inductive ability is enabled through Graph neural networks~\cite{scarselli2008graph} due to their inherent topology-aware setting, which allows the learned system to naturally generalize. Although the idea of graph-based modeling has been suggested for physical systems~\cite{cranmer2020discovering,greydanus2019hamiltonian}, the inductive biases induced due to different graph structures and their consequences on the dynamics remain poorly explored. \rev{In this context, there are a few recent works that aim to develop better representations of physical systems through equivariant graph neural networks~\cite{satorras2021n,han2022learning,huang2021equivariant}. However, the present work focuses on the inductive biases for physical systems rather than focusing on better representation learning. Some of the proposed inductive biases may also be a natural consequence of forcing equivariance. For example, enforcing invariance to the Lagrangian w.r.t. space translation and rotation also enforces Newton's third law, which we achieve through an inductive bias (\S~\ref{sec:ind_bias}). In the context of generalizability to unseen systems, there exists work on \textit{few-shot} based on  \textit{meta-learning}~\cite{metahamiltonian}. Meta-learning assumes the availability of a limited amount of training data to adapt to an unseen system (task). This is different from our objective where we perform zero-shot generalizability, i.e., inference on an unseen system without any further training.} To summarize, our key contributions are as follows:
\begin{enumerate}
\item {\bf Topology-aware modeling}, where the physical system is modeled using a graph-based \node{} (\gnode), \rev{enabling zero-shot generalizability to unseen system sizes}.
\item{\bf Decoupling the dynamics and constraints}, where the equation of motion of the \gnode{} is decoupled into individual terms such as forces, Coriolis-like terms, and explicit constraints.
\item{\bf Decoupling the internal and body forces}, where the internal forces due to the interaction among particles are decoupled from the forces due to external fields.
\item{\bf Newton's third law}, where the forces of interacting particles are enforced to be equal and opposite. We theoretically prove that this inductive bias enforces the conservation of linear momentum, in the absence of an external field, for the predicted trajectory exactly.
\end{enumerate}
We analyze the role of these biases on $n$-pendulum and spring systems. Interestingly, depending on the nature of the system, we show that some of these biases can either significantly improve the performance or have marginal effects. We also show that the final model with the appropriate inductive biases significantly outperforms the simple version of \gnode{} with no additional inductive biases and even the graph versions of \lnn{} and \hnn{}.

\vspace{-0.15in}
\section{Background on dynamical systems}
\vspace{-0.10in}
A dynamical system can be defined by ordinary differential equations (ODEs) of the form, $\ddot{q}=F(q,\dot{q},t)$, where $F$ represents the dynamics of the system. Thus, the time evolution of the system can be obtained by integrating this equation of motion. In Lagrangian mechanics, the dynamics of a system are derived based on the Lagrangian of the system, which is defined as~\cite{goldstein2011classical}:
\begin{equation}
L(q,\dot{q},t)=T(q,\dot{q})-V(q)
\end{equation}
Once the Lagrangian of the system is computed, the dynamics can be obtained using the  Euler-Lagrange (EL) equation as $\frac{d}{dt} \left(\nabla_{\dot{q}}L\right)=\nabla_{q} L$. Note that the EL equation is equivalent to D'Alembert's principle written in terms of the force as $\sum_{i=1}^n (N_i-m_i \ddot q_i)\cdot \delta q_i = 0$, where $\delta q_i$ represents a virtual displacement consistent with the constraints of the system. Thus, an alternative to the EL equation can be written in terms of the forces and other components as~\cite{lavalle2006planning,murray2017mathematical}
\begin{equation}
    M(q) \ddot{q} + C(q,\dot{q}) \dot{q} + N(q) + \Upsilon(\dot{q}) + A^T\lambda = \Pi
\label{eq:node_decoupled}
\end{equation}
where $M(q)$ represents the mass matrix, $C(q,\dot{q})$ the Coriolis-like forces, $N(q)$ the conservative forces, $\Upsilon(\dot q)$ represents the non-conservative forces such as drag or friction, and $\Pi$ represents any external forces. \rev{We note that Lagrangian dynamics is traditionally applied in systems with conservative forces, i.e., where a Lagrangian is well-defined. However, Eq.~\ref{eq:node_decoupled} is generic and can be applied to model any dynamical system irrespective of whether the system has a well-defined Lagrangian or not. Examples of systems where a well-defined Lagrangian may not exist include that of a colloidal system undergoing Brownian motion.} 

The constraints of the system are given by $A(q)\dot{q}=0$, where $A(q) \in \mathbb{R}^{k×D}$ represents $k$ constraints. Thus, $A$ represents the non-normalized basis of constraint forces given by $A=\nabla_q (\phi)$, where $\phi$ is the matrix corresponding to the constraints of the system and $\lambda$ is the Lagrange multiplier corresponding to the relative magnitude of the constraint forces. Note that $M, C,$ and $N$ are directly derivable from the Lagrangian of the system $L$ as $M = \nabla_{\dot q \dot q}L, C = \nabla_{q \dot q} L,$ and $N = \nabla_q L$. Thus, decoupling the dynamic equation simplifies the representation of the dynamics $F$ with enhanced interpretability. To learn the dynamics, Eq.~\ref{eq:node_decoupled} can be represented as 
\begin{equation}
\ddot{q} = M^{-1}\left(\Pi-C\dot{q}-N-\Upsilon-A^T\lambda\right)
\label{eq:acc}
\end{equation}
To solve for $\lambda$, the constraint equation can be differentiated with respect to time to obtain $A\ddot{q}+\dot{A}\dot{q}=0$. Substituting for $\ddot{q}$ from Eq. \ref{eq:acc} and solving for $\lambda$, we get 
\begin{equation}
    \lambda = (A M^{-1}A^T)^{-1}(A M^{-1}(\Pi - C \dot q - N -\Upsilon) + \dot A \dot q)
\end{equation}
Accordingly, $\ddot q$ can be obtained as
\begin{equation}
    \ddot q = M^{-1}\left(\Pi - C\dot q - N - \Upsilon - A^T (A M^{-1}A^T)^{-1}\left(A M^{-1}(\Pi - C \dot q - N -\Upsilon) + \dot A \dot q\right)\right)
    \label{eq:gnode_const}
\end{equation}

\vspace{-0.20in}
\section{Inductive biases for neural ODE}
\vspace{-0.10in}
\label{sec:ind_bias}
To learn dynamical systems, \node{}s parameterize the dynamics $F(q_t,\dot q_t)$ using a neural network and learn the approximate function $\hat{F}(q_t,\dot q_t)$ by minimizing the loss between the predicted and actual trajectories, that is, $\mathcal{L} = ||q_{t+1} - \hat{q}_{t+1}||$. Thus, a \node{} essentially learns a cumulative $\hat{F}$, that consists of the contribution of all the terms contributing to the dynamics. In contrast, in an \lnn, the learned $L$ is used in the EL equation to obtain the dynamics. Similarly, in \lnn with explicit constraints, the learned Lagrangian is used to obtain the individual terms in the Eq.~\ref{eq:gnode_const} by differentiating the \lnn, which is then substituted  to obtain the acceleration. However, this approach is transductive in nature. To overcome this challenge, we present a graph-based version of the \node namely, \gnode{}, which models the physical system as a graph. We show, that \gnode{} can generalize to arbitrary system size after learning the dynamics $\hat{F}$.
\vspace{-0.10in}
\subsection{Graph neural ODE (\gnode) for physical systems}
\vspace{-0.10in}

In this section, we describe the architecture of  \gnode{}. The graph topology is used \rev{to learn} the approximate dynamics $\hat{F}$ by minimizing the loss on the predicted positions. \\
{\bf Graph structure.} An $n$-particle system is represented as an undirected graph $\mathcal{G}=\{\mathcal{V,E}\}$, where nodes represent particles and edges represent the connections or interactions between them. For instance, in a pendulum or springs system, the nodes correspond to bobs or balls, respectively, and the edges correspond to the bars or springs, respectively.\\
{\bf Input features.} Each node is characterized by the features of the particles themselves, namely, the particle \textit{type} ($t$), \textit{position} ($q_i$), and \textit{velocity} ($\dot q_i$). The \textit{type} distinguishes particles of differing characteristics, for instance, balls or bobs with different masses. Further, each edge is represented by the edge features $w_{ij}=(q_i-q_j)$, which represents the relative displacement of the nodes connected by the given edge.\\
\rev{\textbf{Neural architecture:} Here, we provide a summary of the various neural layers within \gnode (see a detailed description in App.~\ref{app:gnode} and ). Fig.~\ref{fig:graph_architecture} in the appendix provides a pictorial description.} \rev{In the \textit{pre-processing} layer, we construct a dense vector representation for each node $v_i$ and edge $e_{ij}$ using MLPs. In many cases, internal forces in a system that govern the dynamics are closely dependent on the topology of the structure. We model this through a \textit{deep, message-passing \textsc{Gnn}}. Let $\cz_i=\ch_i^L$ denote the final representation constructed by the \textsc{Gnn} over each node. These representations are passed through another MLP to predict the acceleration of each node.}\\
\textbf{Trajectory prediction and training.} Based on the predicted $\ddot{q}$, the positions and velocities are predicted using the \textit{velocity Verlet} integration. The loss function of \gnode{} is computed by using the predicted and actual accelerations at timesteps $2, 3,\ldots,\mathcal{T}$ in a trajectory $\mathbb{T}$, which is then back-propagated to train the MLPs. Specifically, the loss function is as follows.
\begin{equation}
    \label{eq:lossfunction}
  \mathcal{L}= \frac{1}{n}\left(\sum_{i=1}^n \rev{\sum_{t=2}^\mathcal{T} }\left(\ddot{q}_i^{\mathbb{T},t}-\hat{\ddot{q}}_i^{\mathbb{T},t}\right)^2\right)
\end{equation}
 Here, $(\hat{\ddot{q}}_i^{\mathbb{T},t})$ is the predicted acceleration for the $i^{th}$ particle in trajectory $\mathbb{T}$ at time $t$ and $\ddot{q}_i^{\mathbb{T},t}$ is the true acceleration. $\mathbb{T}$ denotes a trajectory from $\mathfrak{T}$, the set of training trajectories. Note that the accelerations are computed directly from the ground truth trajectory using the Verlet algorithm:
 \begin{equation}
\ddot{q}(t)=\frac{1}{(\Delta t)^2}[q(t+\Delta t)+q(t-\Delta t)-2q(t)]
\end{equation}
Since the integration of the equations of motion for the predicted trajectory is also performed using the same algorithm as $q(t+\Delta t)=2q(t)-q(t-\Delta t)+\ddot{q}(\Delta t)^2$, equivalent to training from trajectory/positions. It should be noted that the training approach presented here may lead to learning the dynamics as dictated by the Verlet integrator. Thus, the learned dynamics may not represent the ``true'' dynamics of the system, but one that is optimal for the Verlet integrator\footnote{\rev{See App.~\ref{app:gnode} for details on why we use Verlet integrator}}.

\rev{\gnode{} learns Newtonian dynamics directly without decoupling the constraints and other components governing the dynamics from the equation $\ddot q = \hat{F}(q,\dot q, t)$. Next, we analyze the effect of providing the constraints explicitly in the form of inductive biases and enhance the performance of \gnode{}. In the subsequent sections, we iteratively stack inductive biases on top of \gnode{}. Thus, the final proposed architecture, \mcgnode, is the eventual model that considers all biases. The detailed differences between all the graph architectures proposed in the work are included in App.~\ref{app:diff_graph_arch} and Fig.~\ref{fig:all_graph_architecture}.}






\vspace{-0.10in}
\subsection{Decoupling the dynamics and constraints (\cgnode).}
\vspace{-0.10in}
 To this extent, while keeping the graph architecture exactly the same, we use Eq.~\ref{eq:gnode_const} to compute the acceleration of the system. In this case, the output of \gnode{} is $N_i = \texttt{squareplus}(\texttt{MLP}_{\CV}(\cz_i))$ instead of $\ddot{q}_i$, where $N_i$ is the conservative force on particle $i$ (refer Eq.~\ref{eq:node_decoupled}). Then, the acceleration is predicted using Eq.~\ref{eq:gnode_const} with explicit constraints. This approach, termed as \textit{constrained} \gnode{} (\cgnode), enables us to isolate the effect of imposing constraints explicitly on the learning, and the performance of \gnode{}. It is worth noting that for particle systems in Cartesian coordinates, the mass matrix is a diagonal one, with the diagonal entries, $m_{ii}$s, representing the mass of the $i^{th}$ particle. Thus, the $m_{ii}$s are learned as trainable parameters. As a consequence of this design, the graph directly predicts the $N_i$ of each particle. Note that in the case of systems with drag, the graph would contain a cumulative of $N_i$ and $\Upsilon_i$. \rev{Examples of such systems are demonstrated later (see App.~\ref{app:drag_ext_force}).}
\looseness=-1

\begin{wrapfigure}{r}{0.6\columnwidth}
\centering
\includegraphics[width=0.6\columnwidth]{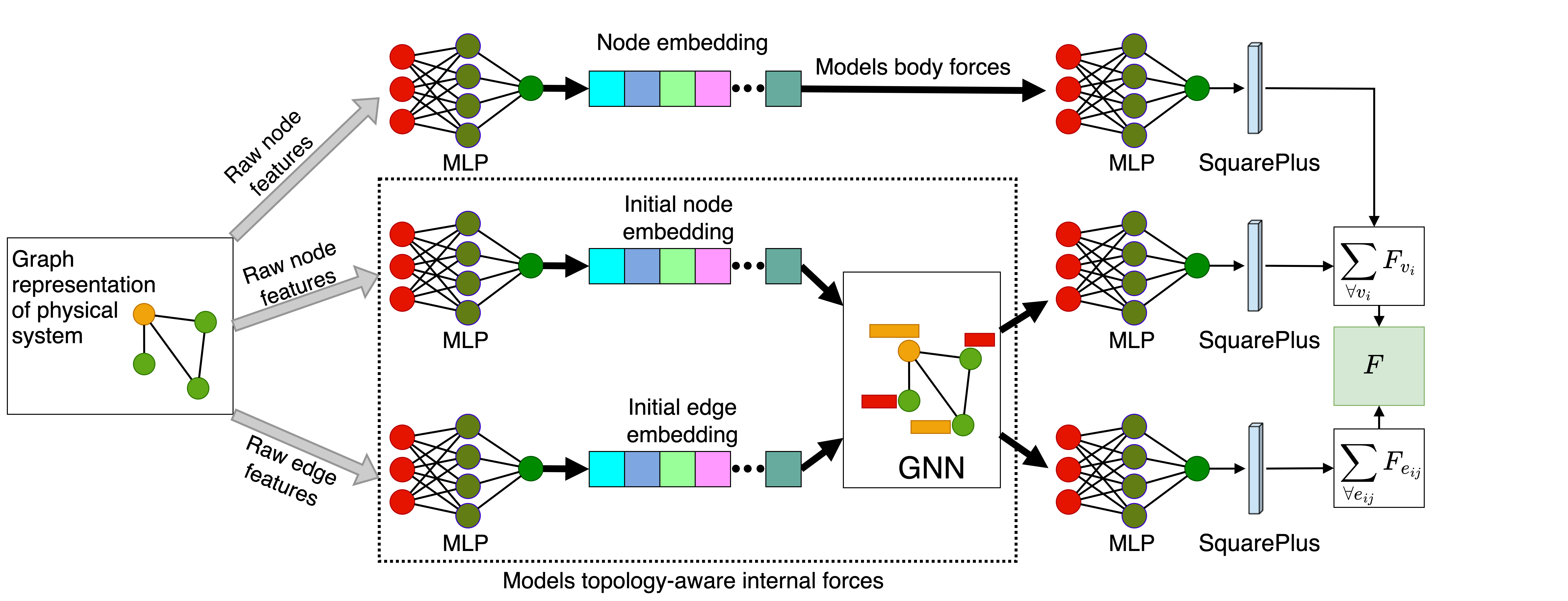}
\vspace{-0.15in}
\caption{Architecture of \cdgnode (see App.~\ref{app:diff_graph_arch}).}\label{fig:cdgnode_arch}
\vspace{-0.20in}
\end{wrapfigure}

\vspace{-0.10in}
\subsection{Decoupling the internal and body forces (\cdgnode)}
\vspace{-0.10in}
In a physical system, internal forces arise due to interactions between constituent particles. On the contrary, the body forces are due to the interaction of each particle with an external field such as gravity or electromagnetic field. While internal forces are closely related to the system's topology, the effect of external fields on each particle is independent of the topology of the structure. In addition, while the internal forces are a function of the relative displacement between the particles, the external forces due to fields depend on the actual position of the particle.

To decouple the internal and body forces, we modify the graph architecture of the \gnode{}. Specifically, the node inputs are divided into local and global features. Only the local features are involved in the message passing. The global features are passed through an MLP to obtain an embedding, which is concatenated with the node embedding obtained after message passing. This updated embedding is passed through an MLP to obtain the particle level force, $N_i$. Here, the local node features are the particle types and the global features are position and velocity. Note that in the case of systems with drag, the $\Upsilon_i$ can be learned using a separate node MLP using particle type and velocity as global input features. This architecture is termed as \textit{constrained and} \textit{decoupled} \gnode{} (\cdgnode). The architecture of \cdgnode~is provided in Fig.~\ref{fig:cdgnode_arch}.

\vspace{-0.10in}
\subsection{Newton's third law (\mcgnode{})} 
\vspace{-0.10in}
According to Newton's third law, \textit{``every action has an equal and opposite reaction''}. In particular, the internal forces on each particle exerted by the neighboring particles should be equal and opposite so that the total system is in equilibrium. That is, $f_{ij}=-f_{ji}$, where $f_{ij}$ represents the force on particle $i$ due to its neighbor $j$. Note that this is a stronger requirement than the EL equation itself, i.e., $\frac{d}{dt} \left(\nabla_{\dot{q_i}}L\right)=\nabla_{q_i} L$. EL equation enforces that the rate of change of momentum of each particle is equal to the total force on the particle, $N_i = \sum_{j=1}^{n_i}f_{ij}$, where $n_i$ is the number of neighbors of $i$ --- it does not enforce any restrictions on the individual component forces, $f_{ij}$ from the neighbors that add to this force. Thus, although the total force on each particle may be learned by the \gnode, the individual components may not be learned correctly. We show that \gnode with an additional inductive bias enforcing Newton's third law, namely \textit{momentum conserving }\gnode{} (\mcgnode), conserves the linear momentum of the learned dynamics exactly.
\vspace{-0.10in}
\begin{thm}
\textit{In the absence of an external field, \mcgnode{} exactly conserves the momentum.}
\label{thm:mom_cons}
\end{thm}
\vspace{-0.10in}
Proof of the theorem is provided in Appendix~\ref{app:proof}. To enforce this bias, we modify the graph architecture as follows. In particular, the output of the graph after the message passing is taken from the edge embedding instead of taking from the node embedding. The final edge embedding is obtained after the message passing is passed through an MLP to obtain the force $f_{ij}$. Then the nodal force is computed as the sum of all the forces from the edges incident on the node. Note that while doing the sum, the symmetry condition $f_{ij}=-f_{ji}$ is enforced. Further, the global forces coming from any external field are added separately to obtain the total nodal forces. Finally, the parameterized masses are learned as the masses in a particle system in Cartesian coordinates are diagonal in nature.

\vspace{-0.20in}
\section{Empirical Evaluation}
\label{sec:experiments}
\vspace{-0.10in}
In this section, we benchmark \gnode{} and the impact of the inductive biases. Our implementation is available at \url{https://anonymous.4open.science/r/graph_neural_ODE-8B3D}. Details of the simulation environment are provided in App.~\ref{app:exp_sys}. 
\vspace{-0.15in}
\subsection{Experimental setup}
\label{sec:setup}
\vspace{-0.10in}
$\bullet$ \textbf{Baselines:} To compare the performance of \gnode{} and its associated versions \mcgnode{}, \cdgnode, and \cgnode, 
 we compare with graph-based physics-informed neural networks, namely, (1) Lagrangian graph network (\lgn)~\cite{lnn}, and (2) Hamiltonian graph network~\cite{sanchez2019hamiltonian}. Since the exact architecture and details of the graph architecture are not provided in the original works, a full graph architecture~\cite{cranmer2020discovering,sanchez2020learning} is used for modeling \lgn~and \hgn\footnote{\rev{We also evaluated the performance of \lgn and \hgn on the \textsc{Gnn} of \gnode and observe that the performance is inferior. Details are provided in App.~\ref{app:gnnlgnhgn}}}. Since Cartesian coordinates are used, the parameterized masses are learned as a diagonal matrix as performed in~\cite{lnn1}. \rev{ We note that while in \gnode{}, the predicted acceleration is the direct output of the graph neural network, in \lgn (or \hgn) the output is the Lagrangian (or Hamiltonian) and further differentiation of the Lagrangian (or Hamiltonian) provides acceleration. These additional computational steps make \lgn or \hgn computationally expensive. We empirically establish this in App.~\ref{app:train_eff}. This result is consistent with earlier results that note neural ODEs (\node{}) as being more efficient and generalizable than Hamiltonian neural networks if appropriate inductive biases are injected \cite{gruver2021deconstructing}. Thus, we also compare with \node{}~\cite{gruver2021deconstructing}. Finally, we also compare with SymODE-Net~\cite{zhong2019symplectic} (App.~\ref{app:symoden}) and graph network simulator (GNS) (App.~\ref{app:gns}).}

\textbf{$\bullet$ Datasets and systems:} To compare the performance \gnode{}s with different inductive biases and with the baseline models, we selected systems on which \node{}s are shown to give good performance~\cite{zhong2021benchmarking,lnn1}, \textit{viz}, $n$-pendulums and springs, where $n=(3,4,5)$. 
Further, to evaluate the zero-shot generalizability of \gnode to large-scale unseen systems, we simulate 5, 20, and 50-link spring systems, and 5-, 10-, and 20-link pendulum systems. The details of the experimental systems are given in Appendix. 
The training data is generated for systems without drag for each particle. The timestep used for the forward simulation of the pendulum system is $10^{-5}s$ with the data collected every 1000 timesteps and for the spring system is $10^{-3}s$ with the data collected every 100 timesteps. The detailed data-generation procedure is given in App.~\ref{app:imple}.

\textbf{$\bullet$ Evaluation Metric:} Following the work of~\cite{lnn1}, we evaluate  performance by computing the following three error metrics, namely, (1) the relative error in  the trajectory, defined as the \textit{rollout error, (RE(t))}, (2) \textit{Energy violation error (EE(t))}, (3) the relative error in \textit{momentum conservation, (ME(t))}, given by: 
$RE(t)=\frac{||\hat{q}(t)-q(t)||_2}{||\hat{q}(t)||_2+||q(t)||_2}; 
EE(t)=\frac{||\hat{\mathcal{H}}-\mathcal{H}||_2}{||\hat{\mathcal{H}}||_2+||\mathcal{H}||_2};
ME(t)=\frac{||\hat{\mathcal{M}}-\mathcal{M}||_2}{||\hat{\mathcal{M}}||_2+||\mathcal{M}||_2}$
Note that all the variables with a hat, for example, $\hat{x}$, represent the predicted values based on the trained model, and the variables without the hat, that is $x$, represent the ground truth.

\textbf{$\bullet$ Model architecture and training setup:} For all versions of \gnode{}, \lgn, and \hgn, the MLPs are two-layers deep. 
We use 10000 data points generated from 100 trajectories to train all the models. This dataset is divided randomly in a 75:25 ratio as a training and validation set. Detailed training procedures and hyper-parameters are provided in App.~\ref{app:imple}. All models were trained till the decrease in loss saturates to less than 0.001 over 100 epochs. The model performance is evaluated on a forward trajectory, a task it was not explicitly trained for, of $10s$ in the case of the pendulum and $20s$ in the case of spring. Note that this trajectory is ~2-3 orders of magnitude larger than the training trajectories from which the training data has been sampled. The dynamics of the $n$-body system are known to be chaotic for $n \geq 2$. Hence, all the results are averaged over trajectories generated from 100 different initial conditions. 
\vspace{-0.15in}
\subsection{Evaluation of inductive biases in \gnode{}}
\vspace{-0.10in}
\begin{figure}
 \vspace{-0.30in}
\centering
\begin{subfigure}{0.49\columnwidth}
    \includegraphics[width=\columnwidth]{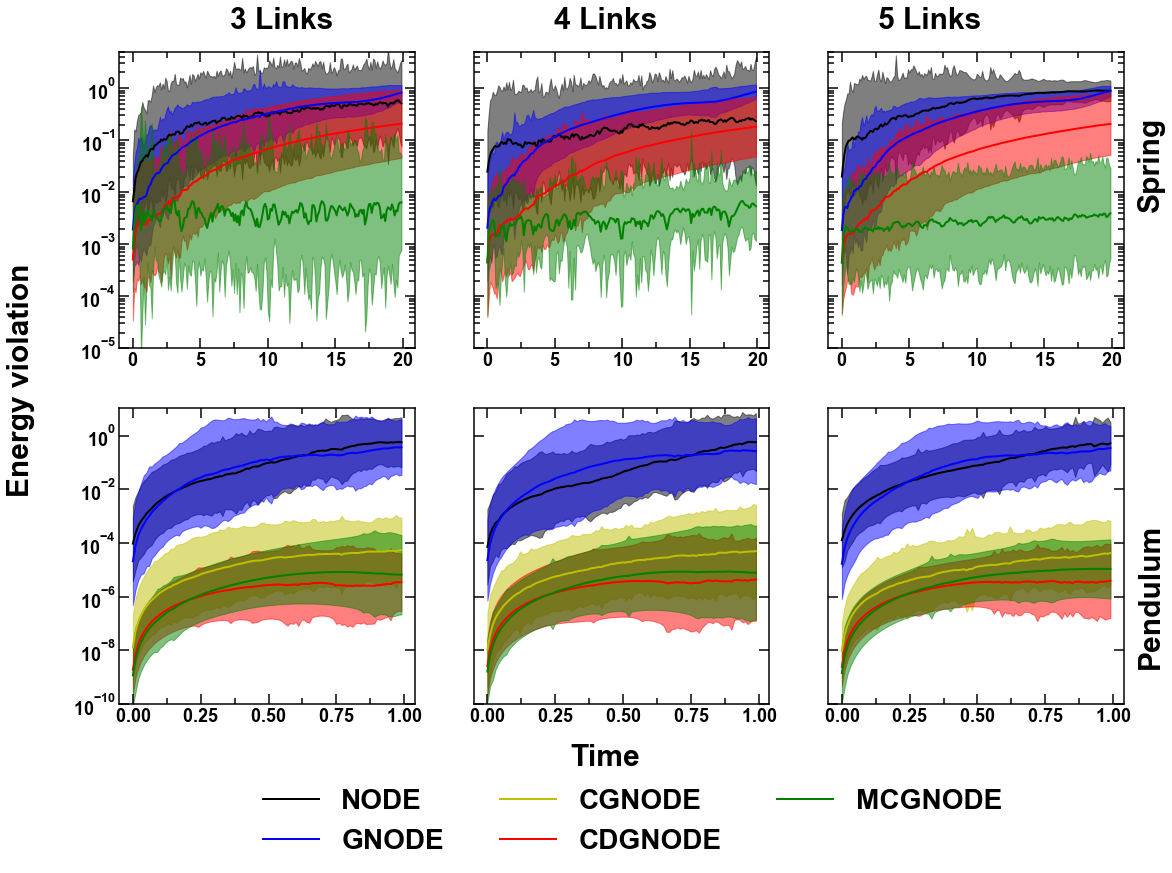}
\end{subfigure}
\begin{subfigure}{0.49\columnwidth}
    \includegraphics[width=\columnwidth]{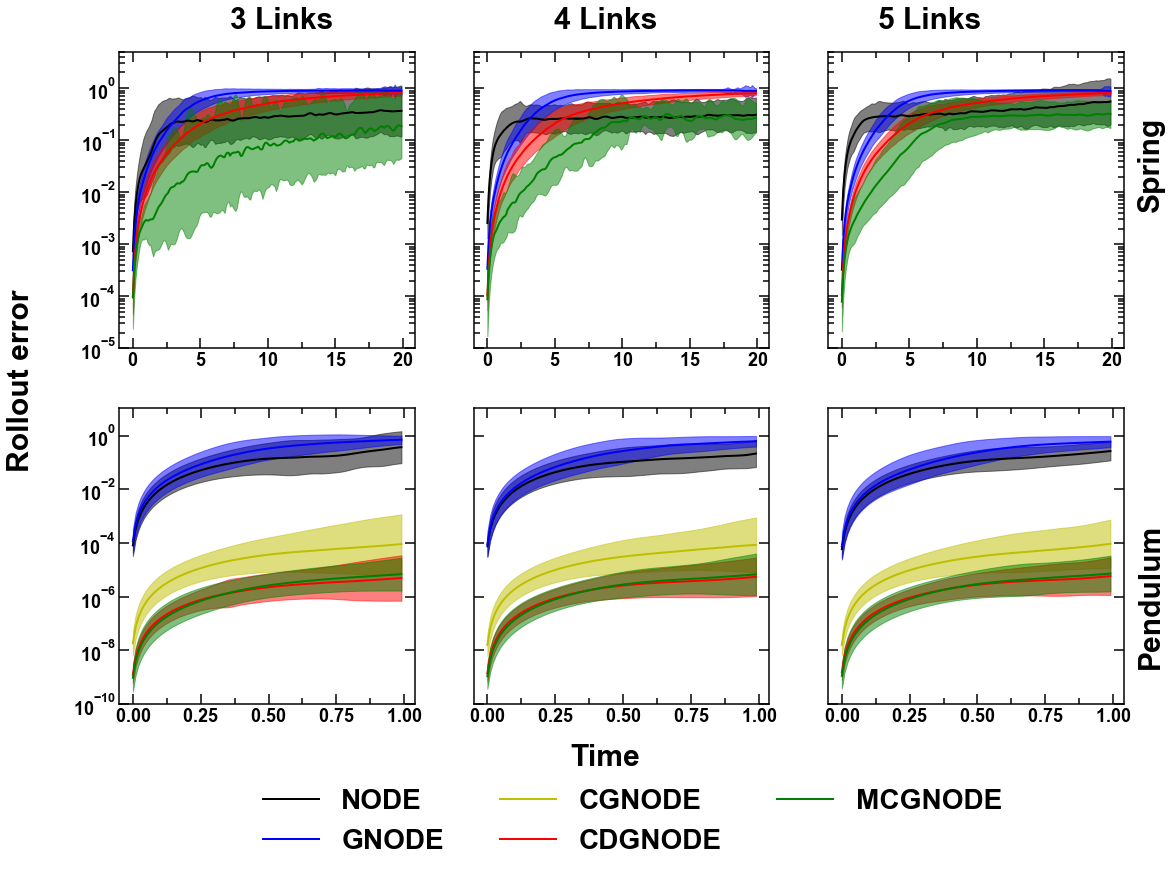}
\end{subfigure}
 \vspace{-0.15in}
\caption{Comparison of the energy violation and rollout error between \node, \gnode, \cgnode, \cdgnode, \mcgnode{} for $n$-pendulum and spring systems with $n=3,4,5$ links. For the pendulum system, the model is trained on the 5-pendulum system and for the spring system, the model is trained on the 5-spring system and evaluated on all other systems. The shaded region shows the 95\% confidence interval from 100 initial conditions.}
\label{fig:fig1}
 \vspace{-0.25in}
\end{figure}

{$\bullet$ \bf Performance of (\gnode).} Fig.~\ref{fig:fig1} shows the performance of \gnode~ on the pendulum and spring systems of sizes 3 to 5. The \gnode for the pendulum is trained on a 5-pendulum system and that of the spring is trained on a 5-spring system. Hence, the 3- and 4-pendulum systems and 3- and 4-spring systems are \textit{unseen}, and consequently, allow us to evaluate zero-shot generalizability. We observe that energy violation and rollout errors of \gnode~for the forward simulations are comparable, both in terms of the value and trend with respect to time, across all of the systems. \rev{Furthermore, when we compare with \node, which has been trained on each system separately, while the errors are comparable in 5-pendulum, \gnode is better in 5-spring. It is worth highlighting that \node has an unfair advantage on 5-pendulum since it is trained and tested on the same system, while 5-pendulum is unseen for \gnode (trained on 5-pendulum). This confirms the generalizability of \gnode to simulate the dynamics of unseen systems of arbitrary sizes.} 
More experiments comparing \node with \gnode to further highlight the benefit of topology-aware modeling are provided in App.\ref{app:topo}.

{$\bullet$ \bf \gnode{} with explicit constraints (\cgnode).} Next, we analyze the effect of incorporating the constraints explicitly in \gnode. To this extent, we train the 5-pendulum system using \cgnode{} model with acceleration computed using Equation~\ref{eq:gnode_const}. Further, we evaluate the performance of the model on 4- and 5-pendulum systems. Note that the spring system is an unconstrained system and hence \cgnode{} is equivalent to \gnode{} for this system. We observe that (see Fig.~\ref{fig:fig1}) \cgnode{} exhibits $\sim$4 orders of magnitude superior performance both in terms of energy violation and rollout error. This confirms that providing explicit constraints significantly improves both the learning efficiency and the predictive dynamics of \gnode{} models where explicit constraints are present, for example, pendulum systems, chains, or rigid bodies.

{$\bullet$ \bf Decoupling internal and body forces (\cdgnode).} Now, we analyze the effect of decoupling local and global features in the \gnode{} by employing the \cdgnode{} architecture. It is worth noting that the decoupling of the local and global features makes the \gnode{} model parsimonious with only the relative displacement and node type given as input features for message passing to the graph architecture. In other words, the number of features given to the graph architecture is lower in \cdgnode{} in \gnode{}. Interestingly, we observe that the \cdgnode{} performs better than \cgnode{} and \gnode{} by $\sim$1 order of magnitude for pendulum and spring systems, respectively.

{$\bullet$ \bf Newton's third law (\mcgnode).} Finally, we empirically evaluate the effect of enforcing Newton's third law, that is, $f_{ij}=-f_{ji}$ on the graph using the graph architecture namely, \mcgnode{}. Figure~\ref{fig:fig1} shows the performance of \mcgnode{} for both pendulum and spring systems. Note that it was theoretically demonstrated that this inductive bias enforces the exact conservation of momentum, in the absence of an external field. Figure~\ref{fig:fig4} shows the $ME$ of \mcgnode in comparison to \node, \gnode, \cgnode, \cdgnode, and \mcgnode for pendulum and spring systems. We observe that in the spring system, the $ME$ of \mcgnode is zero. In the case of a pendulum, although lower than \node, \gnode, and \cgnode, the $ME$ of \mcgnode is comparable to that of \cdgnode. Correspondingly, \mcgnode{} exhibits improved performance for the spring system in comparison to other models. However, in the pendulum system, we observe that the performance of \mcgnode{} is similar to that of \cdgnode{}. This is because, in spring systems, the internal forces between the balls govern the dynamics of the system, whereas in pendulum the dynamics are governed mainly by the force exerted on the bobs by the external gravitational field. Interestingly, we observe that the energy violation error of \mcgnode{} remains fairly constant, while this error diverges for other models. This confirms that strong enforcement of momentum conservation in turn ensures that the energy violation error stays constant over time, ensuring stable long-term equilibrium dynamics of the system. In addition, \mcgnode{} exhibits significantly improved training efficiency than \node{} (see App.\ref{app:train_eff}). \rev{Finally, we also evaluate the ability of \mcgnode{} to learn the dynamics of systems with drag and external forces (see App.~\ref{app:drag_ext_force}). We observe that \mcgnode{} clearly outperforms \node{} in this case as well.} \mcgnode{} will, thus, be particularly useful for the statistical analysis of chaotic systems where multiple divergent trajectories correspond to the same energy state.
\looseness=-1
\vspace{-0.10in}
\subsection{Comparison with baselines}
\vspace{-0.10in}
\begin{figure}[t]
\vspace{-0.30in}
\centering
\begin{subfigure}{0.49\columnwidth}
    \includegraphics[width=\columnwidth]{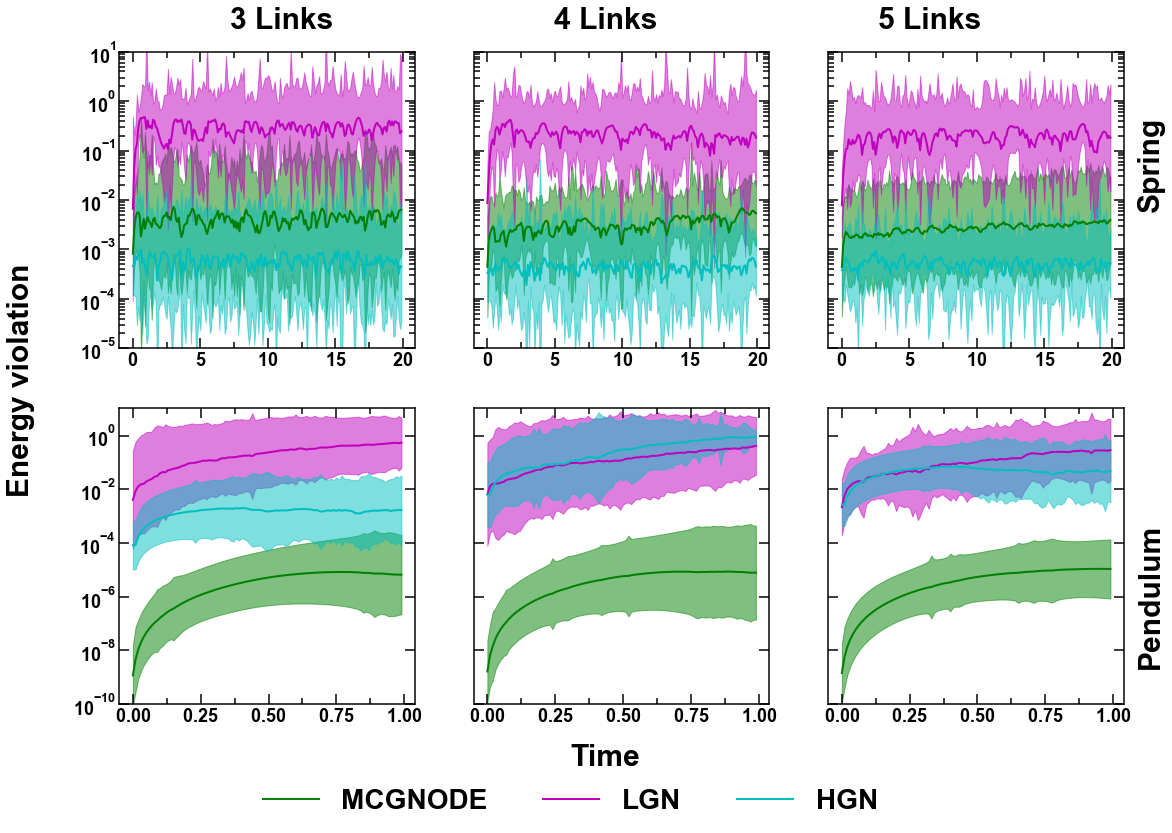}
\end{subfigure}
\begin{subfigure}{0.49\columnwidth}
    \includegraphics[width=\columnwidth]{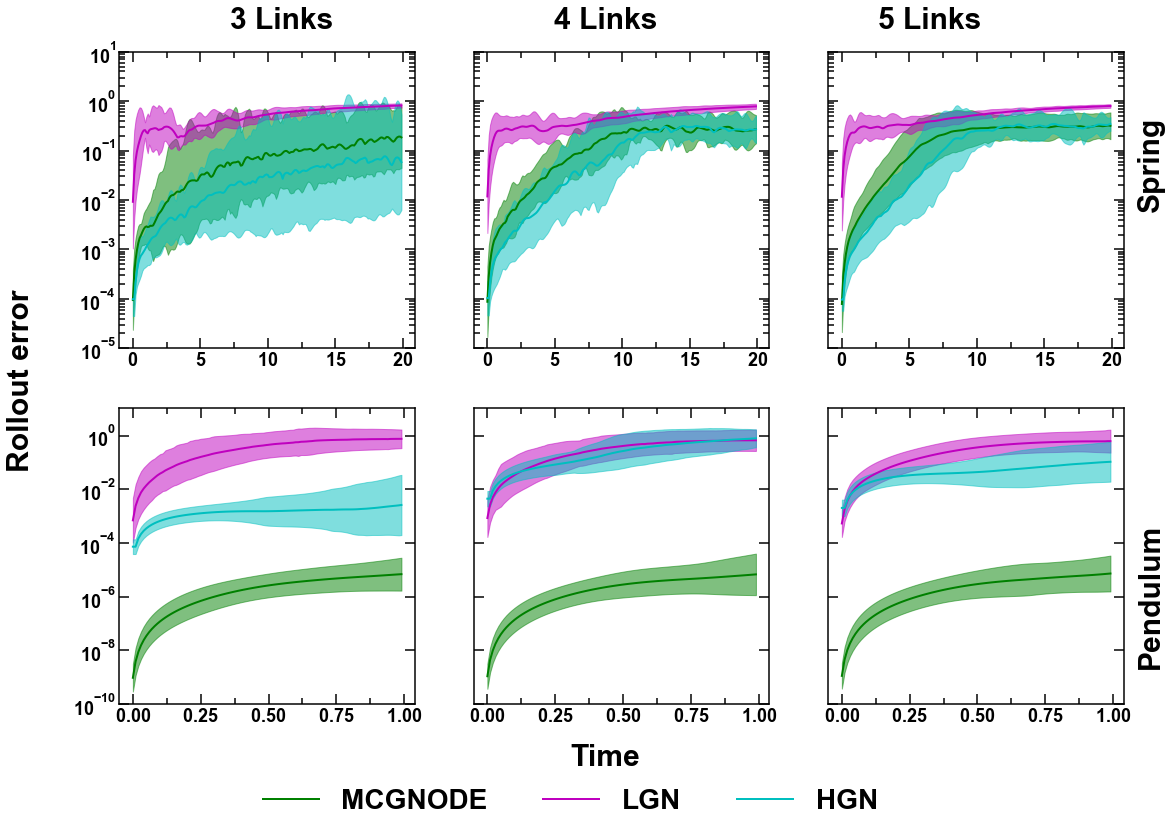}
\end{subfigure}
\vspace{-0.10in}
\caption{Comparison of the energy violation and rollout error between \gnode, \lgn, and \hgn{} for $n$-pendulum and spring systems with $n=3,4,5$ links. For both the pendulum and spring systems, the model is trained on a 5-link structure and evaluated on all other systems. The shaded region shows the 95\% confidence interval from 100 initial conditions.}
\label{fig:fig2}
\vspace{-0.20in}
\end{figure}
\begin{figure}
 \vspace{-0.30in}
\centering
\begin{subfigure}{0.49\columnwidth}
    \includegraphics[width=\columnwidth]{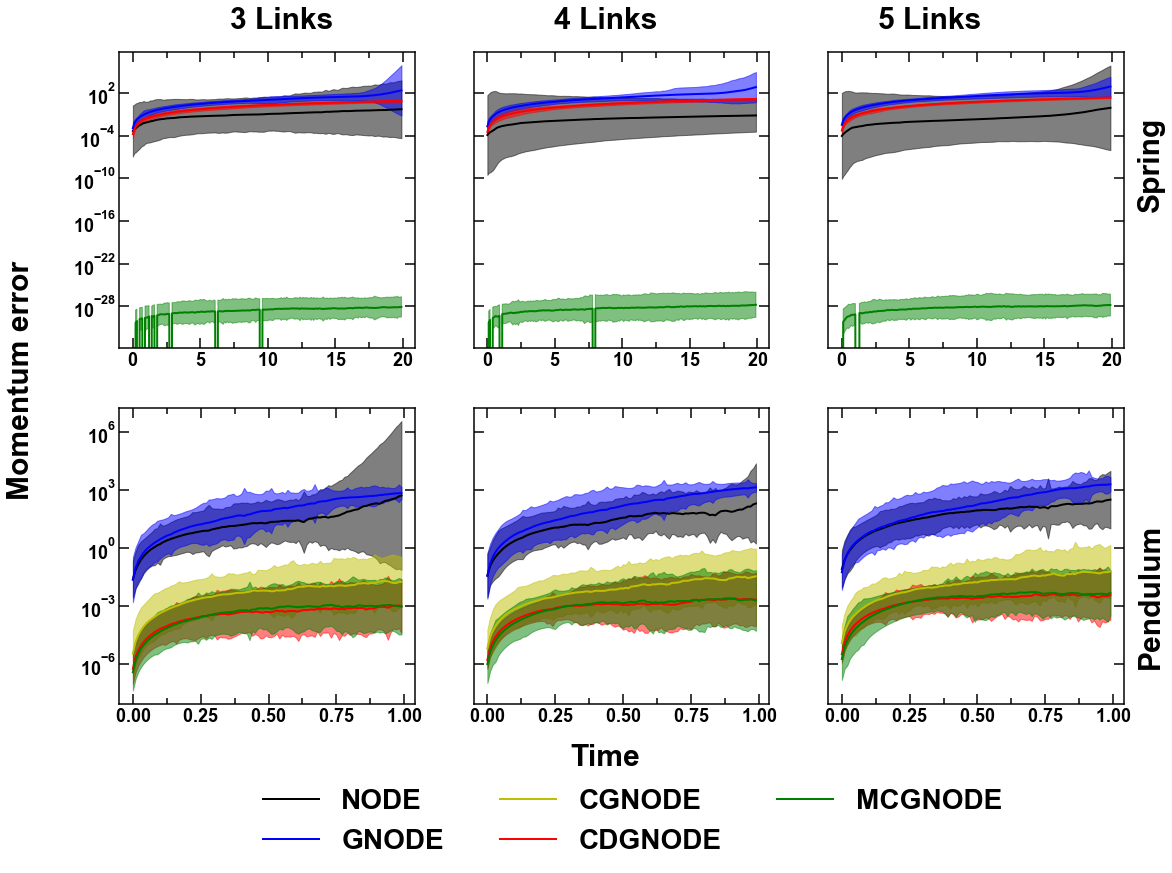}
\end{subfigure}
\begin{subfigure}{0.49\columnwidth}
    \includegraphics[width=\columnwidth]{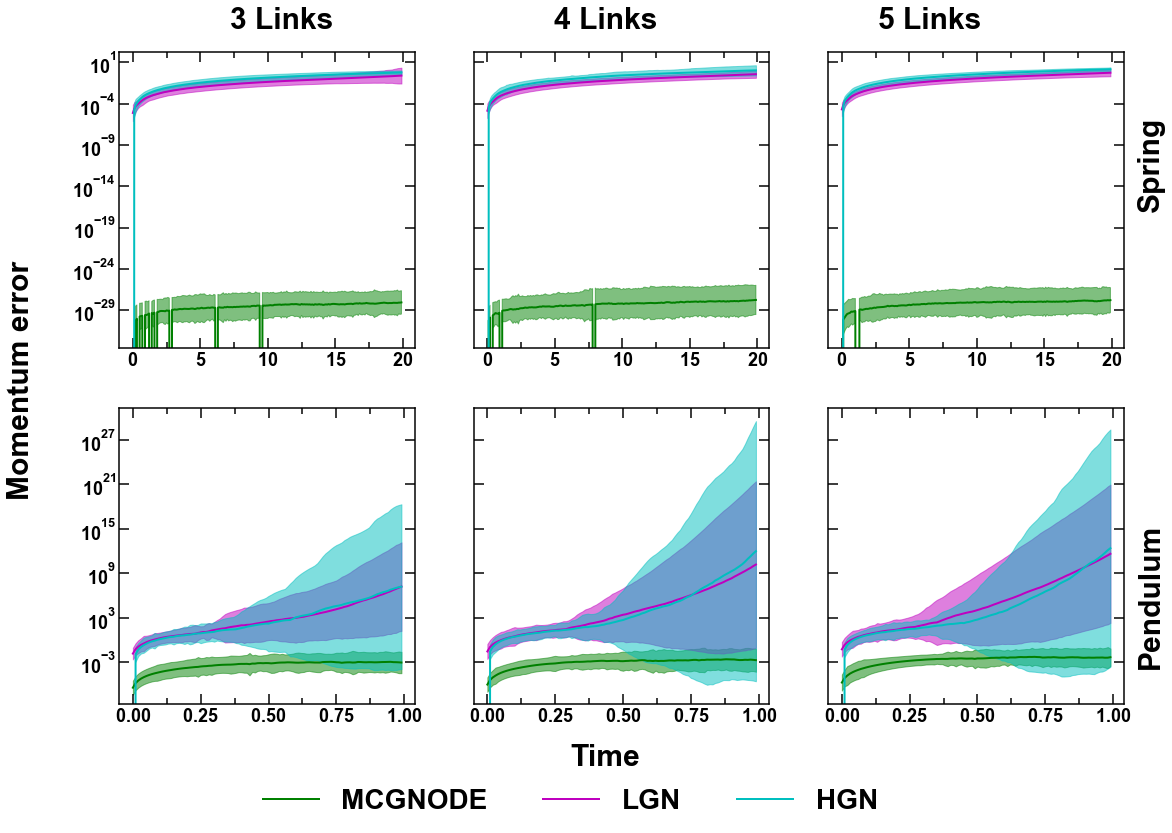}
\end{subfigure}
\vspace{-0.10in}
\caption{Momentum error ($ME$) of \node, \gnode, \cgnode, \cdgnode, and \mcgnode (top palette), and \lgn, \hgn, \mcgnode (bottom palette) for spring (top row of both the palettes) and pendulum (bottom row of both the palettes) systems with 3, 4, and 5 links. The shaded region represents the 95\% confidence interval based on 100 forward simulations.}
\label{fig:fig4}
\vspace{-0.10in}
\end{figure}

Now, we compare the \gnode{} that exhibited the best performance, namely, \mcgnode{} with the baselines namely, \lgn{} and \hgn. Note that both \lgn{} and \hgn{} are energy-conserving graph neural networks, which are expected to give superior performance in predicting the trajectories of dynamical systems~\cite{sanchez2019hamiltonian}. Especially, due to their energy-conserving nature, these models are expected to have lower energy violation error than other models, where energy conservation bias is not explicitly enforced. Figure~\ref{fig:fig2} shows the energy violation and rollout error of \gnode{}, \lgn, and \hgn{} trained on a 5-link system and tested on 3, 4, 5-link pendulum and spring systems. Interestingly, we observe that \mcgnode outperforms both \lgn and \hgn in terms of rollout error and energy error for both spring and pendulum systems. Specifically, in the case of pendulum systems, we observe that the energy violation and rollout errors are $\sim 5$ orders of magnitude lower in the case of \mcgnode in comparison to \lgn and \hgn. Further, we compare the $ME$ of \mcgnode with \lgn and \hgn (see Fig.~\ref{fig:fig4} right palette). We observe in both pendulum and spring systems that the $ME$ of \mcgnode is significantly lower than that of \lgn and \hgn. Thus, in the case of spring systems, although the performance of \hgn{} and \mcgnode are comparable for both $RE$ and $EE$, we observe that \mcgnode exhibits zero $ME$, while in \hgn and \lgn $ME$ diverges. \rev{The superior performance of \mcgnode{} could be attributed to the carefully constructed inductive biases, namely, (i) explicit constraints---improves data efficiency and reduces the error in pendulum systems, (ii) decoupling internal and body forces---leads to superior performance in pendulum systems which act under gravity, (iii) momentum conservation---leads to superior performance in spring systems by enforcing an additional conservation law strongly. We note that these inductive biases may also be adapted suitably for LGN and HGN and thereby improve their performance. Thus, the contribution of our work is not the neural architecture in isolation, but the core design principles that are generic enough to be abstracted beyond neural ODEs.} 
\vspace{-0.10in}
\subsection{Generalizability}
\vspace{-0.10in}
\begin{figure}
\centering
\begin{subfigure}{0.48\columnwidth}
    \includegraphics[width=\columnwidth]{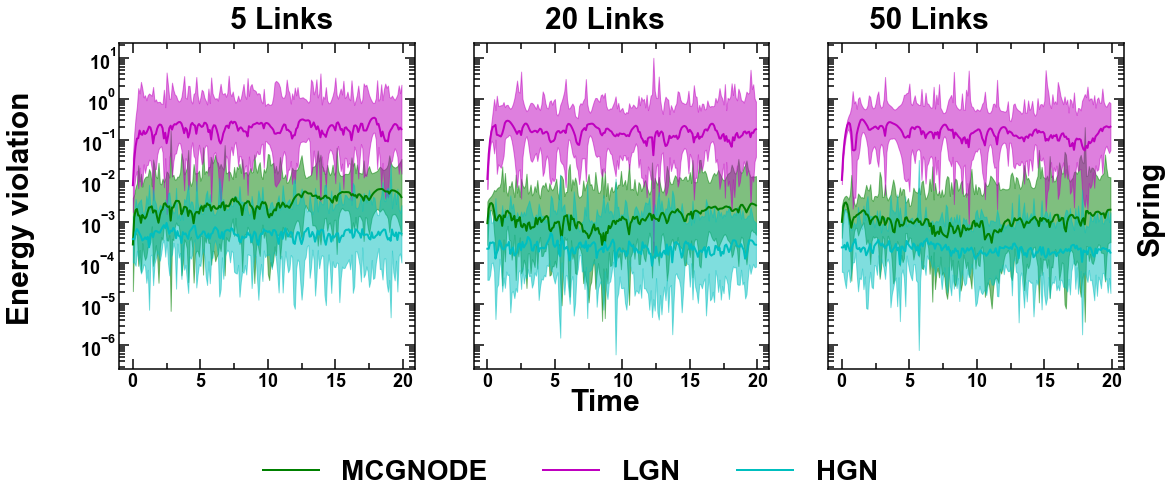}
\end{subfigure}
\begin{subfigure}{0.48\columnwidth}
    \includegraphics[width=\columnwidth]{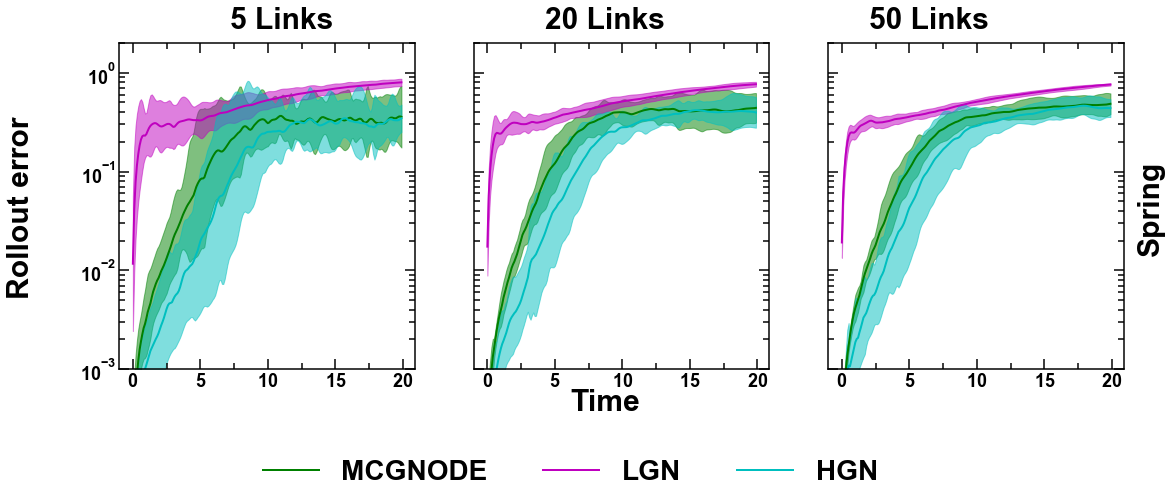}
\end{subfigure}
\begin{subfigure}{0.48\columnwidth}
    \includegraphics[width=\columnwidth]{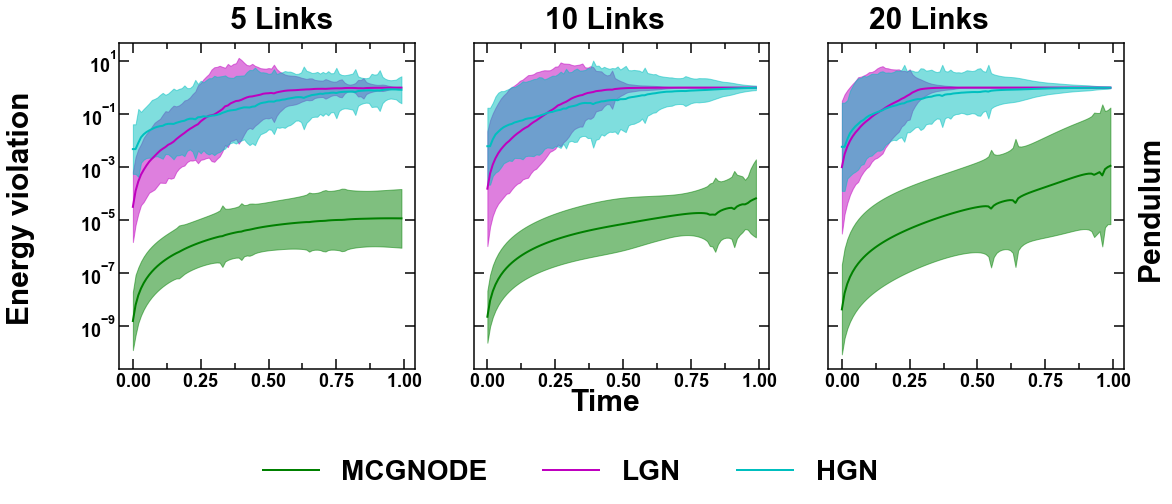}
\end{subfigure}
\begin{subfigure}{0.48\columnwidth}
    \includegraphics[width=\columnwidth]{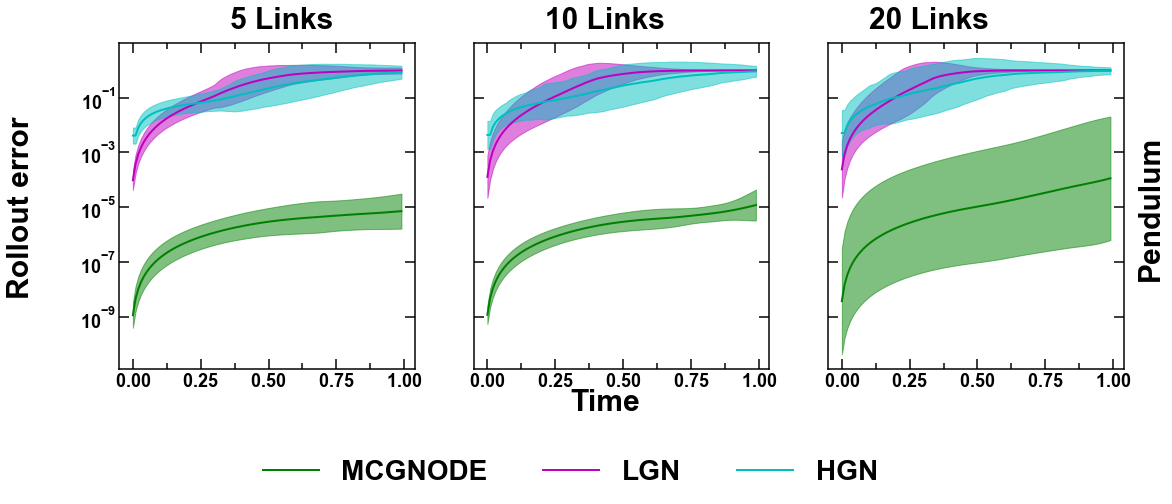}
\end{subfigure}
\vspace{-0.10in}
\caption{Energy violation and rollout error between \gnode, \lgn, and \hgn{} for spring systems with $5, 20, 50$ links and pendulum system with $5, 10, 20$ links. Note that for all the systems, \gnode{}, \lgn, and \hgn are trained on 5-pendulum and 5-spring systems and tested on others. The shaded region represents the $95\%$ confidence interval based on $100$ forward simulations.}
\label{fig:fig3}
\vspace{-0.20in}
\end{figure}
Now, we focus on the zero-shot generalizability of \gnode, \lgn, and \hgn to large systems that are unseen by the model. To this extent, we model 5-, 10-, and 20-link pendulum systems and 5-, 20-, and 50-link spring systems. Note that all the models are trained on the 5-link systems. In congruence with the comparison earlier, we observe that \mcgnode performs better than both \lgn and \hgn when tested on systems with a large number of links. However, we observe that the variation in the error represented by the 95\% confidence interval increases with an increase in the number of particles. This suggests that the increase in the number of particles essentially results in increased noise in the prediction of the forces. However, the error in the \mcgnode is significantly lower than the error in \lgn and \hgn for both pendulum and spring systems, as confirmed by the 95\% confidence interval\rev{, which could again be attributed to the three inductive biases infused in \mcgnode{}, that are absent in \lgn{} and \hgn.} This suggests that \mcgnode can generalize to larger system sizes with increased accuracy.

\vspace{-0.10in}
\subsection{Learning the dynamics of 3D solid}
\vspace{-0.10in}
\begin{wrapfigure}{r}{0.6\textwidth}
\centering
\includegraphics[width=0.58\textwidth]{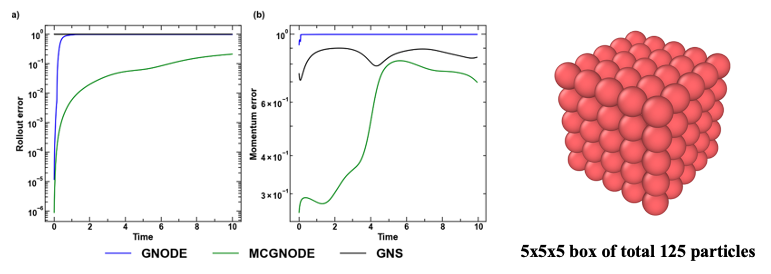}
\vspace{-0.1in}
\caption{($RE$) and ($ME$) of \gnode, \mcgnode and \rev{GNS} for particle-based system. Simulation is performed using peridynamics on a 3D box of $125$ particles ($5 \times 5 \times 5$).\label{fig:fig9}} 
\vspace{-0.20in}
\end{wrapfigure}
To demonstrate the applicability of \mcgnode to learn the dynamics of complex systems, we consider a challenging problem, that is, inferring the dynamics of an elastically deformable body. Specifically, a $5 \times 5 \times 5$ solid cube, discretized into $125$ particles, is used for simulating the ground truth. A 3D solid system is simulated using the peridynamics framework. The system is compressed isotropically and then released to simulate the dynamics of the solid, that is, the contraction and expansion in 3D. We observe that \mcgnode performs inference much superior to \gnode and GNS (see Fig.~\ref{fig:fig9}\rev{and App.~\ref{app:gns}}). It should also be noted that although the momentum error in \mcgnode seems high, this is due to the fact that the actual momentum in these systems is significantly low. This is demonstrated by plotting the absolute momentum error in these systems which in \mcgnode is close to zero, while in \gnode it explodes (see Fig.~\ref{fig:abs_peri_gnode_vs_mcgnode}). These results suggest that the \mcgnode could be applied to much more complex systems such as solids or other particle-based systems where momentum conservation is equally important as energy conservation.
\vspace{-0.20in}
\subsection{\rev{Robustness to Noise}}
\label{sec:noise}
\begin{wrapfigure}{r}{0.4\textwidth}
\vspace{-0.5in}
\centering
\includegraphics[width=0.4\textwidth]{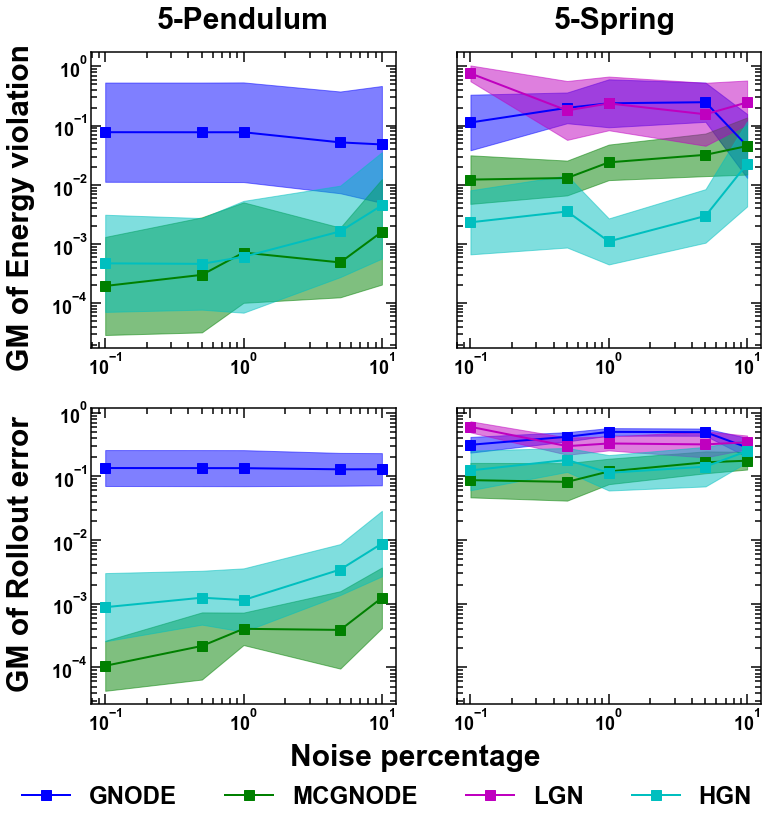}
\vspace{-0.20in}
\caption{\rev{Impact of noise on performance of \gnode{}, \mcgnode, \lgn, and \hgn. Note that the noise is given in terms of \% standard deviation of the distribution of the original output values.\label{fig:fig_noise}}}
\vspace{-0.20in}
\end{wrapfigure}

\vspace{-0.10in}
\rev{Finally, we show the robustness to noise of the \mcgnode{} in comparison to \gnode, and additional baselines, namely, \hgn{} and \lgn{}. Fig.~\ref{fig:fig_noise} shows the performance of these models on 5-spring and 5-pendulum systems for varying noise values. Note that the noise is given in terms of \% standard deviation of the distribution of the original output values. We observe that \mcgnode is robust against noise in comparison to other models.}

\vspace{-0.15in}
\section{Conclusions}
\label{sec:conclusion}
\vspace{-0.15in}
In this work, we carefully analyze various inductive biases that can affect the performance of a \gnode. Specifically, we analyze the effect of a graph architecture, explicit constraints, decoupling global and local features that makes the model parsimonious, and Newton's third law that enforces a stronger constraint on the conservation of momentum. We observe that inducing appropriate inductive biases results in an enhancement in the performance of \gnode by $\sim$2-3 orders magnitude for spring systems and $\sim$5-6 orders magnitude for pendulum systems in terms of energy violation error. Further, \mcgnode outperforms the graph versions of both Lagrangian and Hamiltonian neural networks, namely, \lgn and \hgn. Altogether, the present work suggests that the performance of \gnode-based models can be enhanced and even made superior to energy-conserving neural networks by inducing the appropriate inductive biases. \\
\textbf{Limitations and Future Works:} The present work focuses on particle-based systems such as pendulum and spring systems. While these simple systems can be used to evaluate model performances, real life is more complex in nature with rigid body motions, contacts, and collisions. Extending \gnode to address these problems can enable the modeling of realistic systems. A more complex set of systems involve plastically deformable systems such as steel rod, and viscoelastic materials where the deformation depends on the strain rate of materials. \rev{Further, instead of explicitly providing the constraints, learning the constraints directly from the trajectory can enhance the application of the models to unseen systems. Note that the present work is limited to physical systems. A similar approach could be employed in other dynamical systems as well.} From the neural modeling perspective, the \gnode is essentially a graph convolutional network. Using \textit{attention} over each message passing head may allow us to better differentiate neighbors based on the strength of their interactions. We hope to pursue these questions in the future.

\bibliographystyle{unsrt}
\bibliography{references}


\newpage
\appendix

\begin{figure}
\centering
\includegraphics[width=\columnwidth]{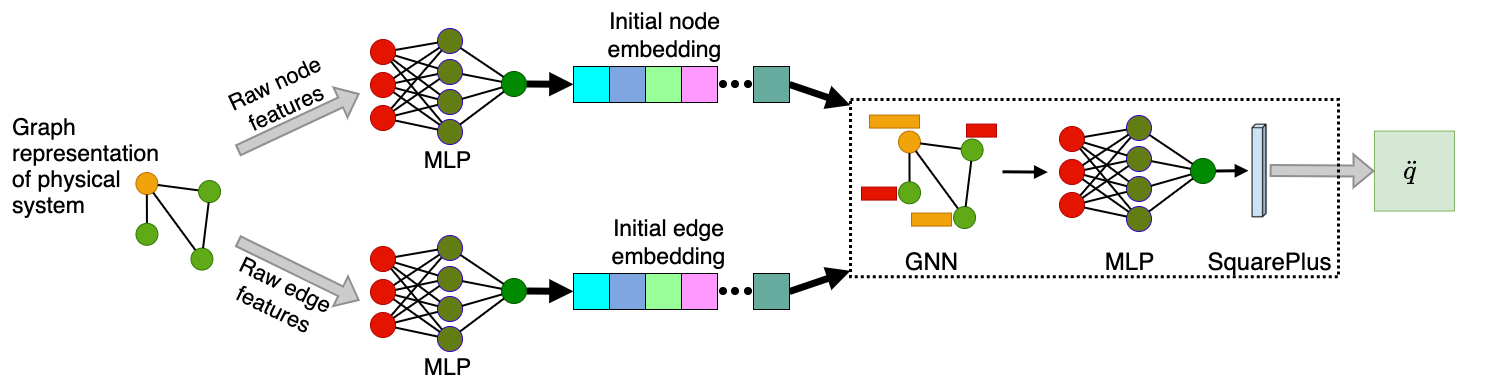}
\caption{Architecture of \gnode.}
\label{fig:graph_architecture}
\end{figure}
\section{Details of \gnode}
\label{app:gnode}
\rev{
The architecture of \gnode is depicted in Fig.~\ref{fig:graph_architecture}. \gnode constitutes of the following neural layers.
\textbf{Pre-Processing.} In the pre-processing layer, we construct a dense vector representation for each node $v_i$ and edge $e_{ij}$ using $\texttt{MLP}_{em}$ as:
\begin{alignat}{2}
    \ch^0_i &= \sqp(\MLP_{em}(\texttt{one-hot}(t_i),q_i,\dot{q}_i)) \label{eq:one-hot}\\
    \ch^0_{ij} &= \sqp(\MLP_{em}(w_{ij}))
\end{alignat}
$\sqp$ is an activation function. Note that the $\MLP_{em}$ corresponding to the node and edge embedding functions are parameterized with different weights. Here, for the sake of brevity, we simply mention them as $\MLP_{em}$.\\
\textbf{Acceleration prediction.} In many cases, internal forces in a system that govern the dynamics are closely dependent on the topology of the structure. To capture this information, we employ multiple layers of \textit{message-passing} between the nodes and edges. In the $l^{th}$ layer of message passing, the node embedding is updated as:
\begin{equation}
    \ch_i^{l+1} = \texttt{squareplus} \left( \ch_i^{l}+\sum_{j \in \mathcal{N}_i}\cW_{\CV}^l\cdot\left(\ch_j^l || \ch_{ij}^l\right) \right)
\end{equation}
where, $\mathcal{N}_i=\{v_j\in\CV\mid e_{ij}\in\CE \}$ are the neighbors of $v_i$. $\cW_{\CV}^{l}$ is a layer-specific learnable weight matrix.
$\ch_{ij}^l$ represents the embedding of incoming edge $e_{ij}$ on $v_i$ in the $l^{th}$ layer, which is computed as follows.
\begin{equation}
    \ch_{ij}^{l+1} = \texttt{squareplus} \left( \ch_{ij}^{l} + \cW_{\CE}^{l}\cdot\left(\ch_i^l || \ch_{j}^l\right) \right)
\end{equation}
Similar to $\cW_{\CV}^{l}$, $\cW_{\CE}^{l}$ is a layer-specific learnable weight matrix specific to the edge set. The message passing is performed over $L$ layers, where $L$ is a hyper-parameter. The final node and edge representations in the $L^{th}$ layer are denoted as $\cz_i=\ch_i^L$ and $\cz_{ij}=\ch_{ij}^L$ respectively. Finally, the acceleration of the particle $\ddot q_i$ is predicted as:
\begin{equation}
\ddot q_i=\texttt{squareplus}(\texttt{MLP}_{\CV}(\cz_i))
\end{equation}
}

\rev{\textbf{Why Verlet integrator:} The canonical coordinates employed in Newtonian mechanics are the position ($q$) and velocity ($\dot{q}$) (Sor momentum) as these quantities are the direct observables. Verlet algorithm is obtained by substracting the Taylor series expansion of $r(t+\Delta t)$ with $r(t-\Delta t)$. The expression thus obtained is time-reversible and has error to the order of ($\mathcal{O}(\Delta t^4)$). Due to the time-reversible nature, Verlet algorithm is energy conserving (symplectic), and hence performs better than other traditional integrators such as RK4 in predicting the long term trajectory of dynamical systems without drift.}
\section{Proof of Theorem \ref{thm:mom_cons}}
\label{app:proof}
\textsc{Proof.} Consider a system without any external field where the dynamics is governed by internal forces, for instance, a $n$-spring system. The conservation of linear momentum for this system implies that $\sum_{i=1}^n \frac{d}{dt} \left(\hat{m}_{ii}\hat{\dot{q}}_i\right) = 0 $, where $\hat{\dot{q}}_i$ represents the predicted velocity, $\hat{m}_{ii}$ represents the learned mass of the particle $i$, and $\dot{q}_i$ represents the ground truth velocity. Assume the error in the rate of change of momentum of the predicted trajectory to be $\hat \varepsilon_{t}$. To prove that \mcgnode{} conserves the momentum exactly, it is sufficient to prove that $\hat \varepsilon_{t} = 0$.
Consider,
\begin{align}
    \hat \varepsilon_{t} &= \sum_{i=1}^n \hat \varepsilon_{i,t} = \sum_{i=1}^n \frac{d}{dt} \left(\hat{m}_{ii}\hat{\dot{q}}_i - m_{ii}\dot{q}_i \right) = \sum_{i=1}^n(\hat{N}_i - N_i) =  \sum_{i=1}^n\hat{N}_i - \sum_{i=1}^nN_i
\end{align}
Note that in equation \ref{eq:acc}, all terms  except $N_i$ are either known or are zero. Thus, without loss of generality, $\sum_{i=1}^nN_i = 0$, in the absence of an external field. Hence,
\begin{align}
    \hat \varepsilon_{t} &= \sum_{i=1}^n \hat \varepsilon_{i,t} = \sum_{i=1}^n\hat{N}_i 
    =\sum_{i=1}^n \sum_{j=1}^{n_i} f_{ij} 
    =\sum_{i=1}^n \sum_{j=1}^{n_i} \frac{(f_{ij}+f_{ji})}{2}
\end{align}
The inductive bias $f_{ij}=-f_{ji}$ implies that $\frac{(f_{ij}+f_{ji})}{2} = 0$. Thus, 
\begin{equation}
\nonumber
    \sum_{i=1}^n \frac{d}{dt} \left(m_{ii}\hat{\dot{q}}_i\right) = \hat \varepsilon_{t} = 0 \hspace{1in}\square
\end{equation} 

\section{\rev{Performance of \lgn and \hgn when implement on the \textsc{Gnn} of \gnode.}}
\label{app:gnnlgnhgn}

\begin{figure}
\centering
\begin{subfigure}{0.49\columnwidth}
    \includegraphics[width=\columnwidth]{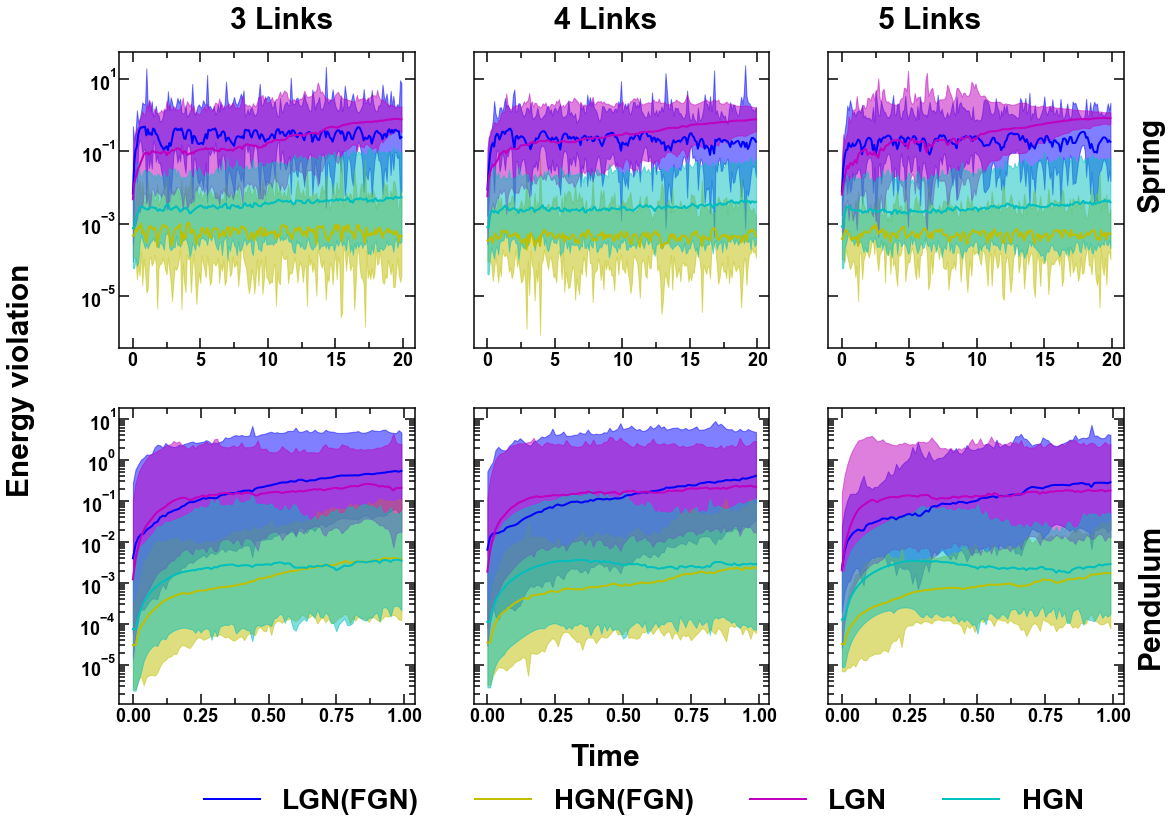}
\end{subfigure}
\begin{subfigure}{0.49\columnwidth}
    \includegraphics[width=\columnwidth]{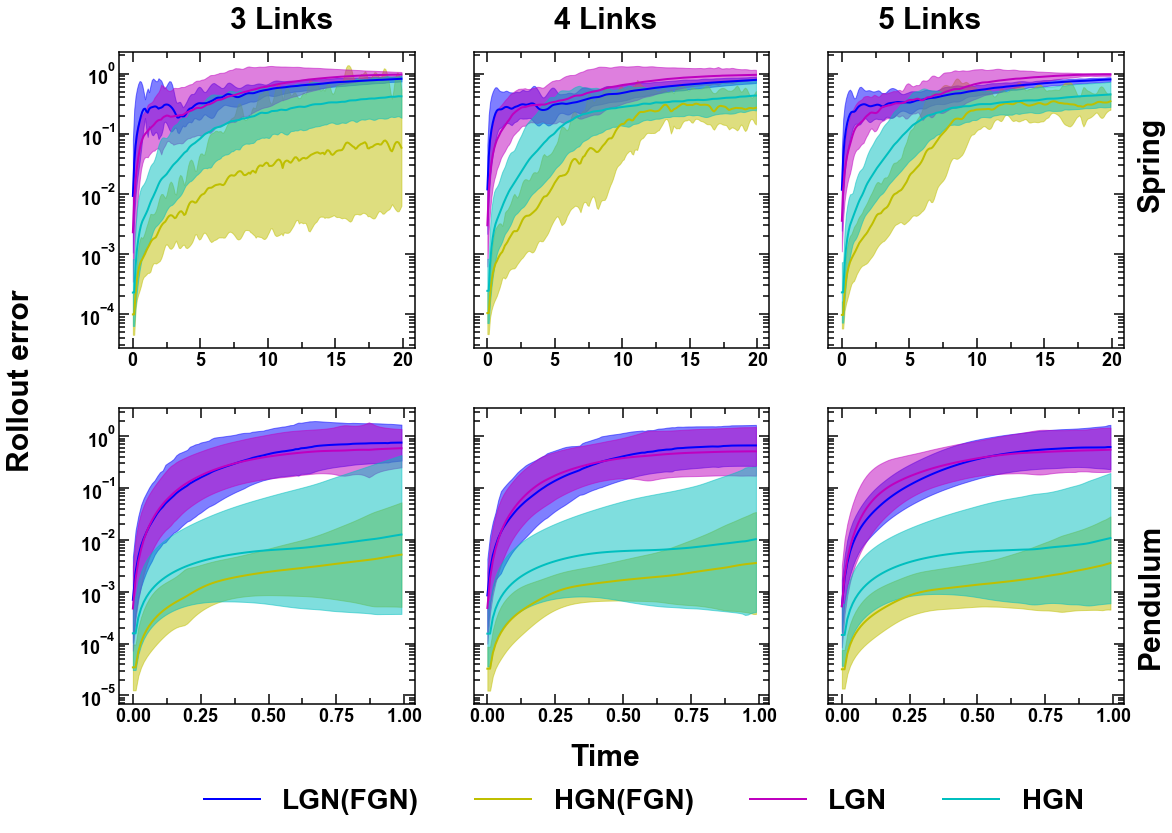}
\end{subfigure}
\vspace{-0.10in}
\caption{\rev{comparison of \lgn and \hgn with the full graph architecture.\label{fig:gnnlgnhgn}}}
\vspace{-0.20in}
\end{figure}

\rev{In Fig.~\ref{fig:gnnlgnhgn}, we compare the performance of \lgn and \hgn implemented on the full graph network~\cite{cranmer2020discovering,sanchez2020learning} against the versions implemented on the \textsc{Gnn} used in \gnode. As visible, the full graph network version provides superior performance.}

\section{\rev{Zero-shot generalizability:} \node vs \gnode vs \mcgnode}
\label{app:topo}
\begin{figure}
\centering
\begin{subfigure}{0.49\columnwidth}
    \includegraphics[width=\columnwidth]{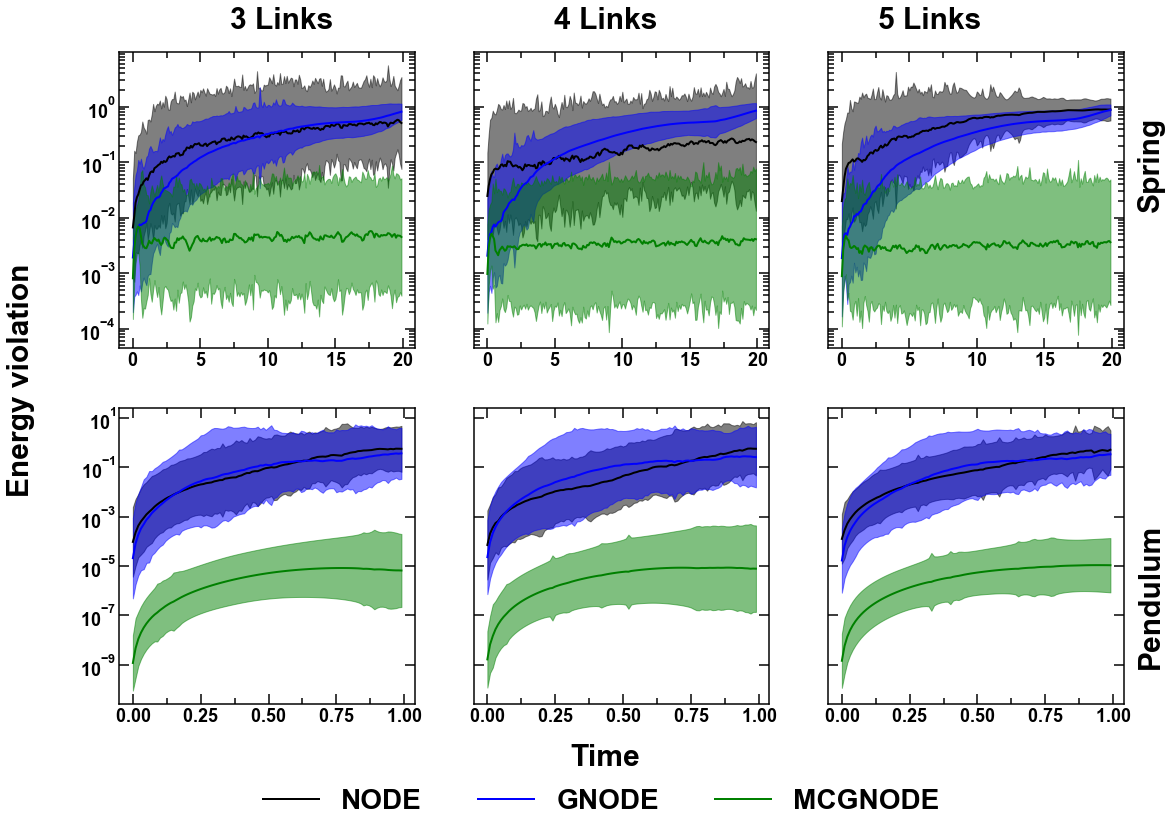}
\end{subfigure}
\begin{subfigure}{0.49\columnwidth}
    \includegraphics[width=\columnwidth]{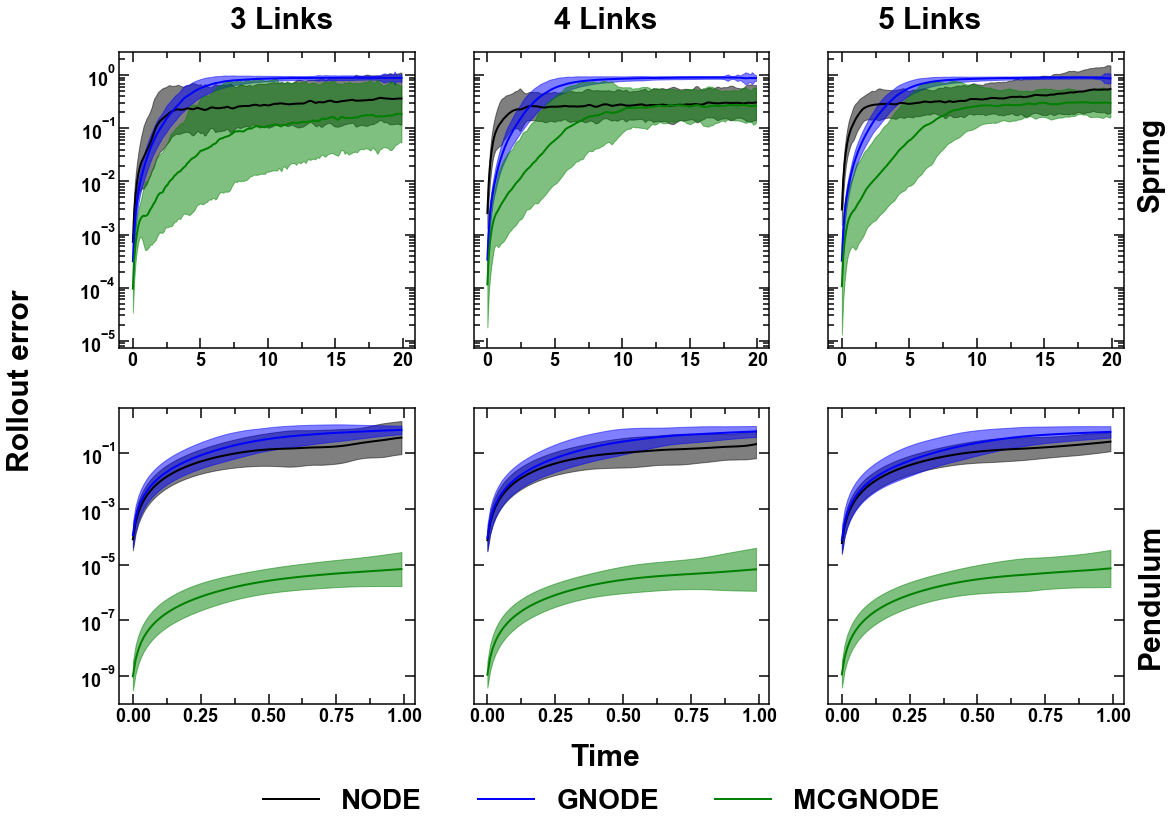}
\end{subfigure}
\caption{Energy error and trajectory error for \node, \gnode, and \mcgnode for spring and pendulum systems with 3-,4-, and 5-links. The shaded region represents the 95\% confidence interval based on 100 forward simulations.}
\label{fig:fig6}
\end{figure}

\rev{To showcase zero-shot generalizability}, we compare the performance of \node, \gnode, \mcgnode in Figure~\ref{fig:fig6}. In the case of \node, the system is trained and evaluated on each of the pendulum and spring systems separately. In the case of \gnode and \mcgnode, the training is performed on 5-spring and 5-pendulum system and evaluated on all other systems. First, we observe that the performance of \node and \gnode are comparable for both rollout and energy error. This suggests that the performance of \gnode, with no additional inductive bias, is comparable to that of \node with the additional advantage that \gnode can generalize to any system size. Now, we compare the performance of \node with \mcgnode, which has all the additional inductive biases discussed earlier. We observe that \mcgnode performs significantly better than \node, as expected. It is worth noting that while the inductive bias of explicit constraints can be infused in \node in a similar fashion as in \gnode, other inductive biases such as decoupling the global and local features and Newton's third law are unique to the graph structure. These inductive biases cannot be infused in a \node structure. This, in turn, demonstrates the superiority of \gnode which allows additional inductive biases due to the explicit representation of the topology through the graph structure.

\begin{figure}
\centering
\begin{subfigure}{0.49\columnwidth}
    \includegraphics[width=\columnwidth]{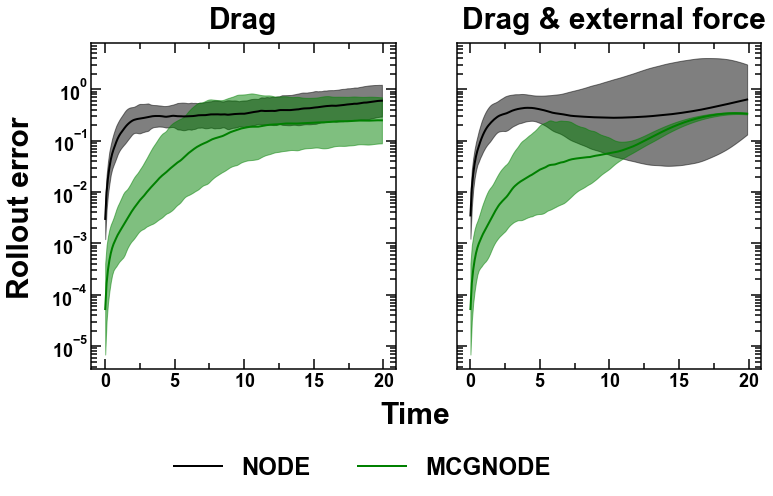}
\end{subfigure}
\begin{subfigure}{0.49\columnwidth}
    \includegraphics[width=\columnwidth]{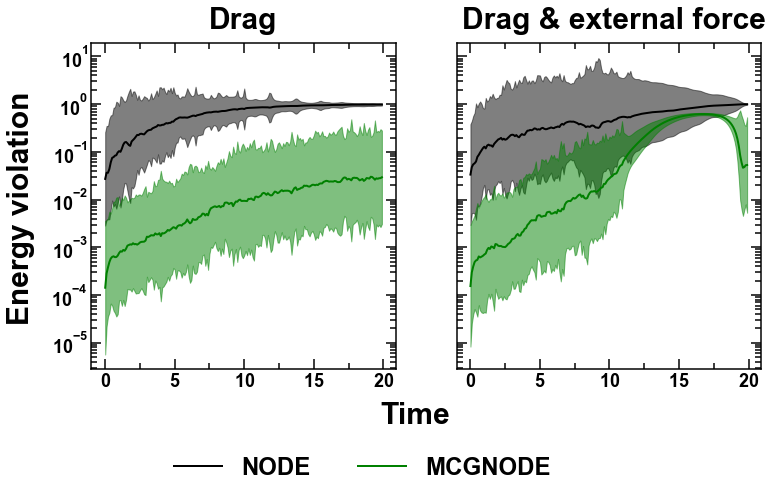}
\end{subfigure}
\vspace{-0.10in}
\caption{\rev{\mcgnode on 5-link spring with non-zero drag and external force with \node as baseline.\label{fig:fig_FY}}}
\vspace{-0.20in}
\end{figure}

\rev{\section{Drag and external force}
\label{app:drag_ext_force}
Here, we evaluate the ability of \mcgnode{} to learn the dynamics of systems subjected to drag linear in velocity (drag force given by $-0.1\dot{q}$) and an external force. Fig.~\ref{fig:fig_FY} shows the performance of \mcgnode{} and \node{} on a 5-spring system with linear drag and external force. We observe that the \mcgnode{} exhibits superior performance in both the cases.}

\section{Training efficiency}
\label{app:train_eff}
\begin{figure}[htp]
\centering
\includegraphics[width=0.65\columnwidth]{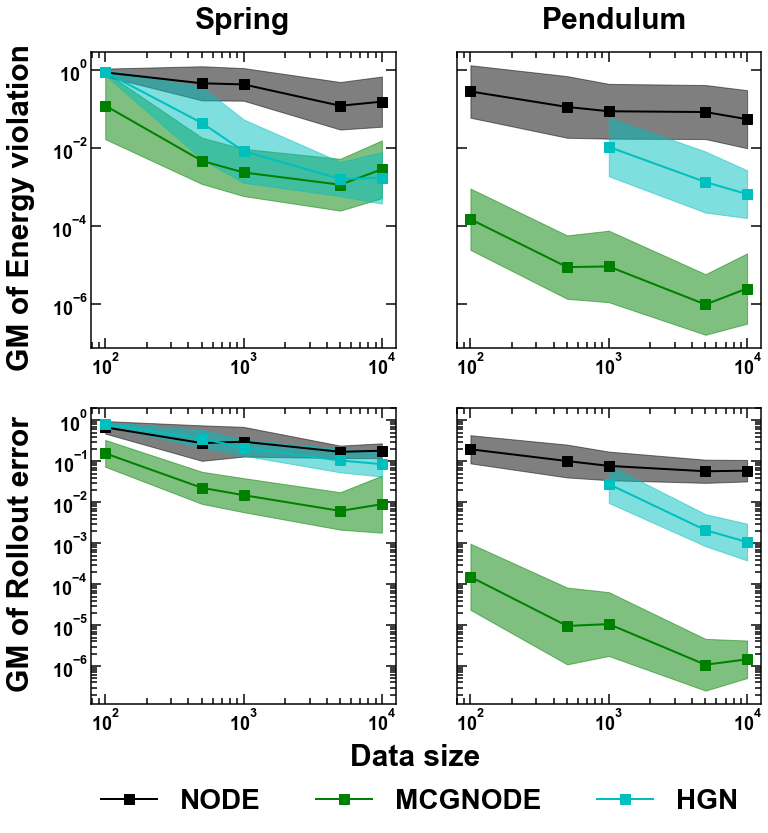}
\caption{Training efficiency of \mcgnode in comparison with the \gnode. The shaded region represents the 95\% confidence interval based on 100 forward simulations.}
\label{fig:fig5}
\end{figure}

{\bf Data Efficiency: }Here, we compare the training efficiency of \mcgnode with the \node \rev{and \hgn. Despite our best efforts, \lgn{} was unable to learn the dynamics from the small amount of data provided and was exploding up on forward simulation leading to NaN values. Similar behavior was observed on \hgn{} with low values data points}. We focus on the best graph-based architecture and compare the performance with the baseline \node 5-spring and 5-pendulum systems. Figure~\ref{fig:fig5} shows the performance of \mcgnode in comparison to \node for the dataset sizes of 100, 500, 1000, 5000, and 10000 in terms of the geometric mean of the energy error and rollout error. We observe that the learning of \mcgnode is significantly simplified in comparison to that of \node---\mcgnode exhibits orders of magnitude better performance in terms of both energy and rollout error for both pendulum and spring systems. \rev{Further, we observe that \mcgnode{} performs orders of magnitude better than \hgn{}}.
\begin{table}[t!]
\centering
{\scriptsize
\begin{tabular}{ |l|c|c| }
\toprule
\textbf{Models} & \textbf{Training time (in sec)} & \textbf{Forward Simulation time (in sec)} \\
\midrule
\mcgnode & 2433 & 0.87\\
\gnode & 6341 & 1.02\\
\hgn & 2512 & 0.76\\
\lgn & 55042 & 14.82\\
\bottomrule
\end{tabular}}
\caption{\rev{Training and inference times in pendulum systems.}}
\label{tab:pendulumtime}
\end{table}

\begin{table}[t!]
\centering
{\scriptsize
\begin{tabular}{ |l|c|c| }
\toprule
\textbf{Models} & \textbf{Training time (in sec)} & \textbf{Forward Simulation time (in sec)} \\
\midrule
\mcgnode & 1576 & 0.06\\
\gnode & 6977 & 0.10\\
\hgn & 14053 & 0.43\\
\lgn & 141962 & 5.81\\
\bottomrule
\end{tabular}}
\caption{\rev{Training and inference times in spring systems.}}
\label{tab:springtime}
\end{table}
\rev{{\bf Computational Efficiency: } We further compare the computation times of the training and inference phases on \gnode, \mcgnode, \hgn and \lgn. Tables~\ref{tab:pendulumtime}-\ref{tab:springtime} present the results. As visible, \mcgnode, is the fastest on average. The difference with \lgn and \hgn is the highest in spring systems. \mcgnode benefits from the inductive biases as well as the fact that the neural architecture directly predicts the force (acceleration is computed from this) on each particle. In contrast, \hgn and \lgn predict the Hamiltonian and Langrangian, which needs to be further differentiated to obtain acceleration. \lgn is the slowest since it requires double differentiation.}

\section{Errors on Internal and External Forces}
\label{app:forceerrors}
\begin{figure}
\centering
\includegraphics[width=0.99\columnwidth]{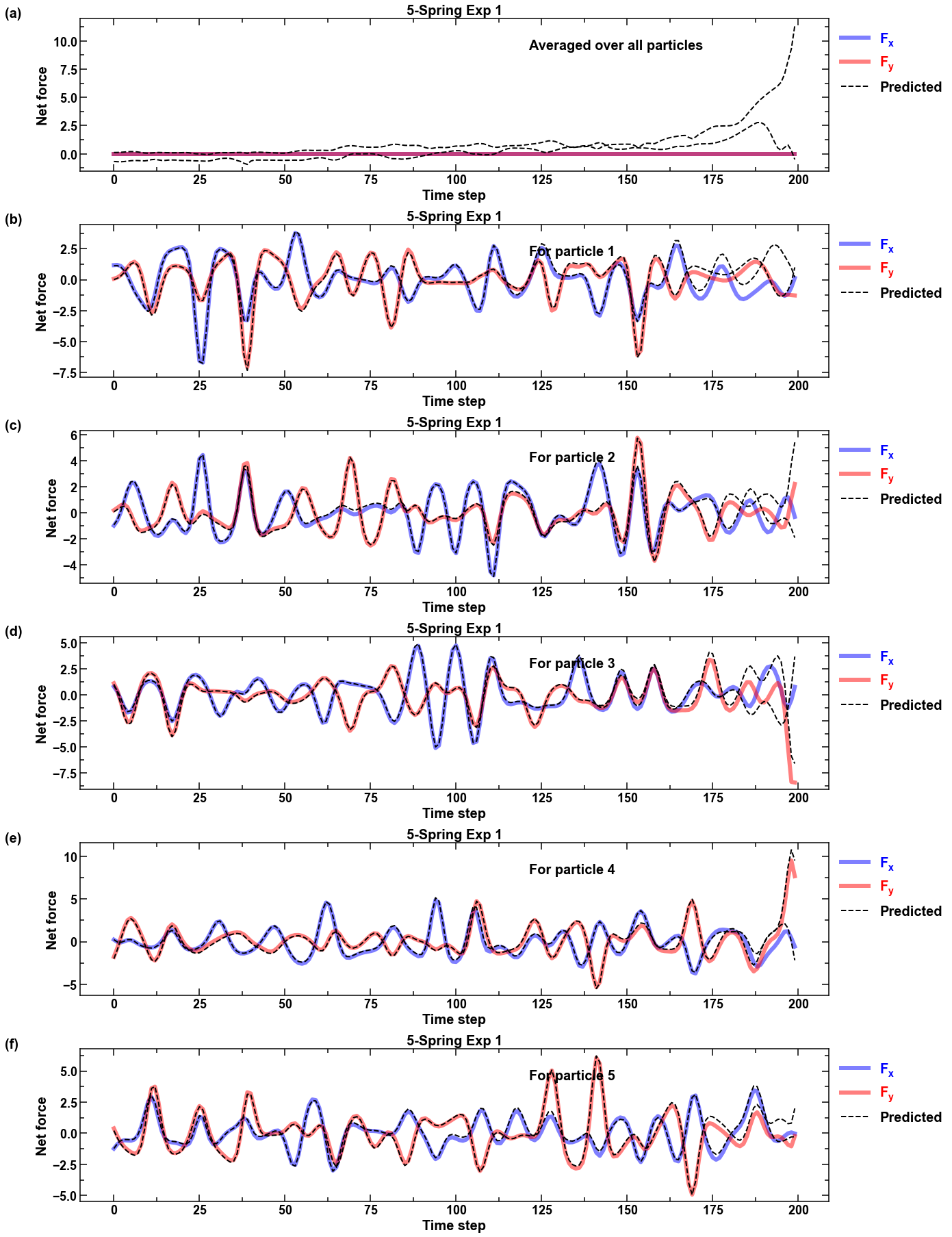}
\caption{\rev{Error in force calculation using \gnode. The solid lines represent the true force and the dashed lines represent the predicted ones. A close overlap among the solid and dashed lines indicate low error. $F_x$ and $F_y$ denote the total forces in the $x$ and $y$ direction respectively. $F_z$ is always 0.}}
\label{fig:fig_c}
\end{figure}

\begin{figure}
\centering
\includegraphics[width=0.99\columnwidth]{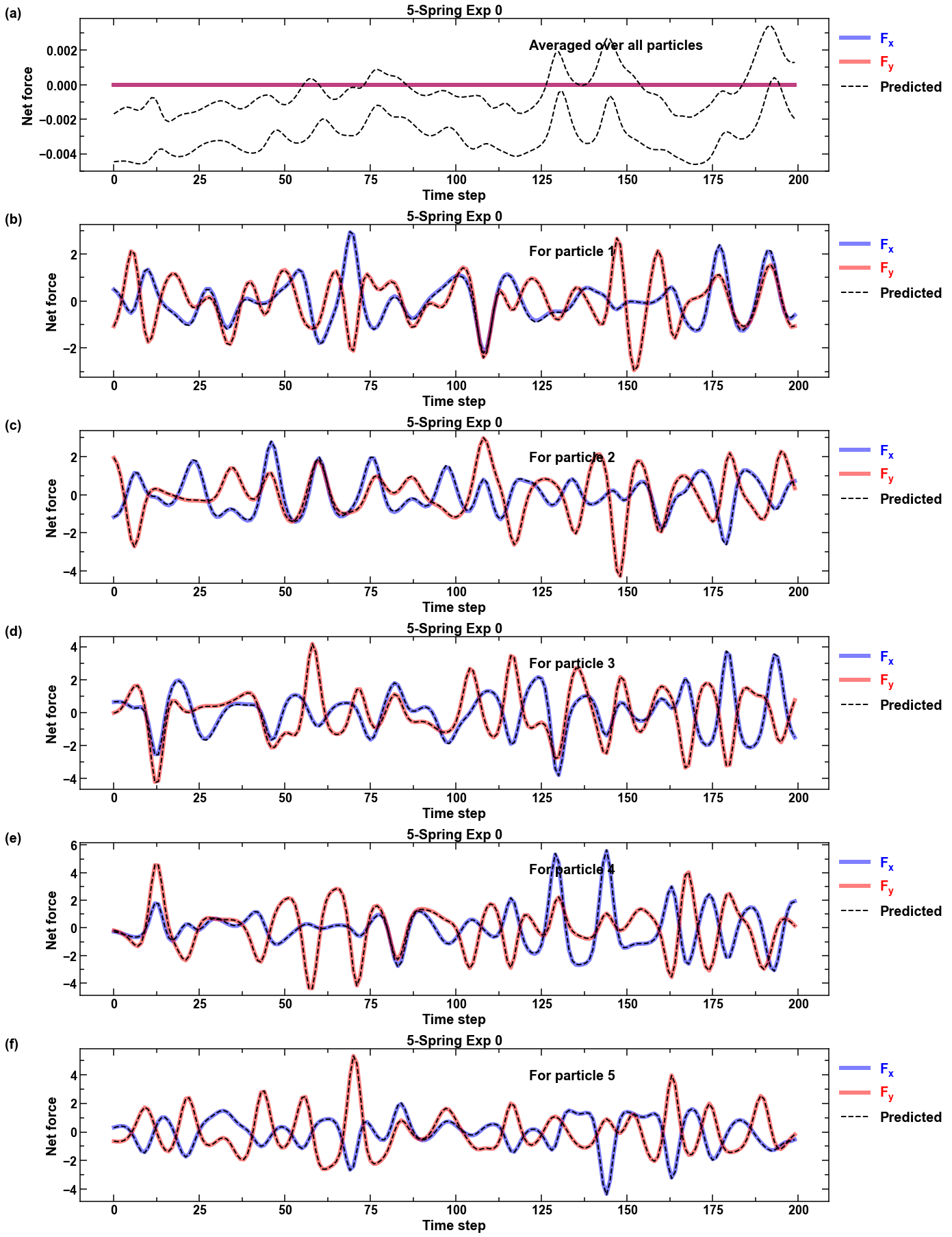}
\caption{\rev{Error in force calculation using \mcgnode. The solid lines represent the true force and the dashed lines represent the predicted ones. A close overlap among the solid and dashed lines indicate low error. $F_x$ and $F_y$ denote the total forces in the $x$ and $y$ direction respectively. $F_z$ is always 0.}}
\label{fig:fig_d}
\end{figure}

\begin{figure}
\centering
\includegraphics[width=0.99\columnwidth]{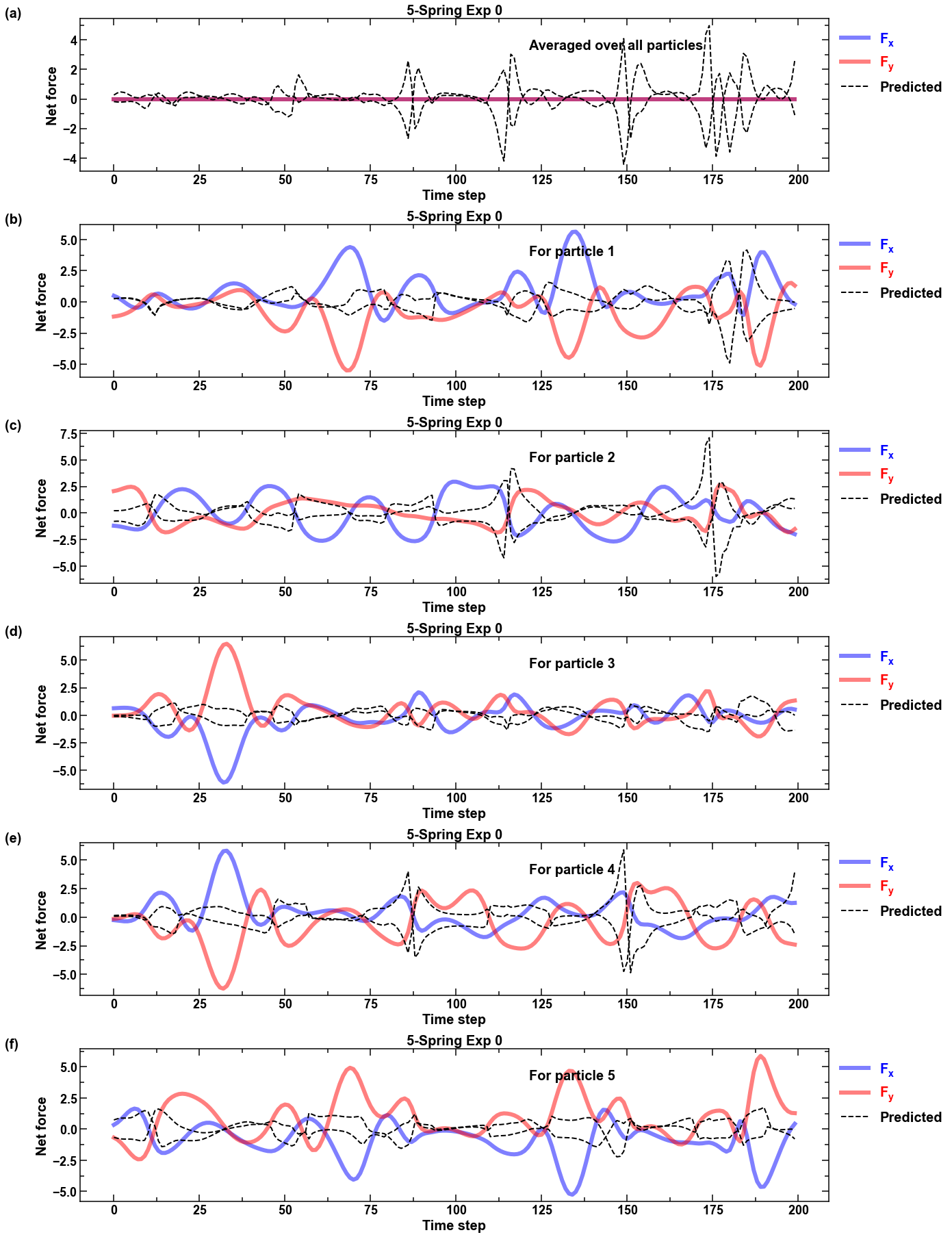}
\caption{\rev{Error in force calculation using \lgn. The solid lines represent the true force and the dashed lines represent the predicted ones. A close overlap among the solid and dashed lines indicate low error. $F_x$ and $F_y$ denote the total forces in the $x$ and $y$ direction respectively. $F_z$ is always 0.}}
\label{fig:fig_a}
\end{figure}

\begin{figure}
\centering
\includegraphics[width=0.99\columnwidth]{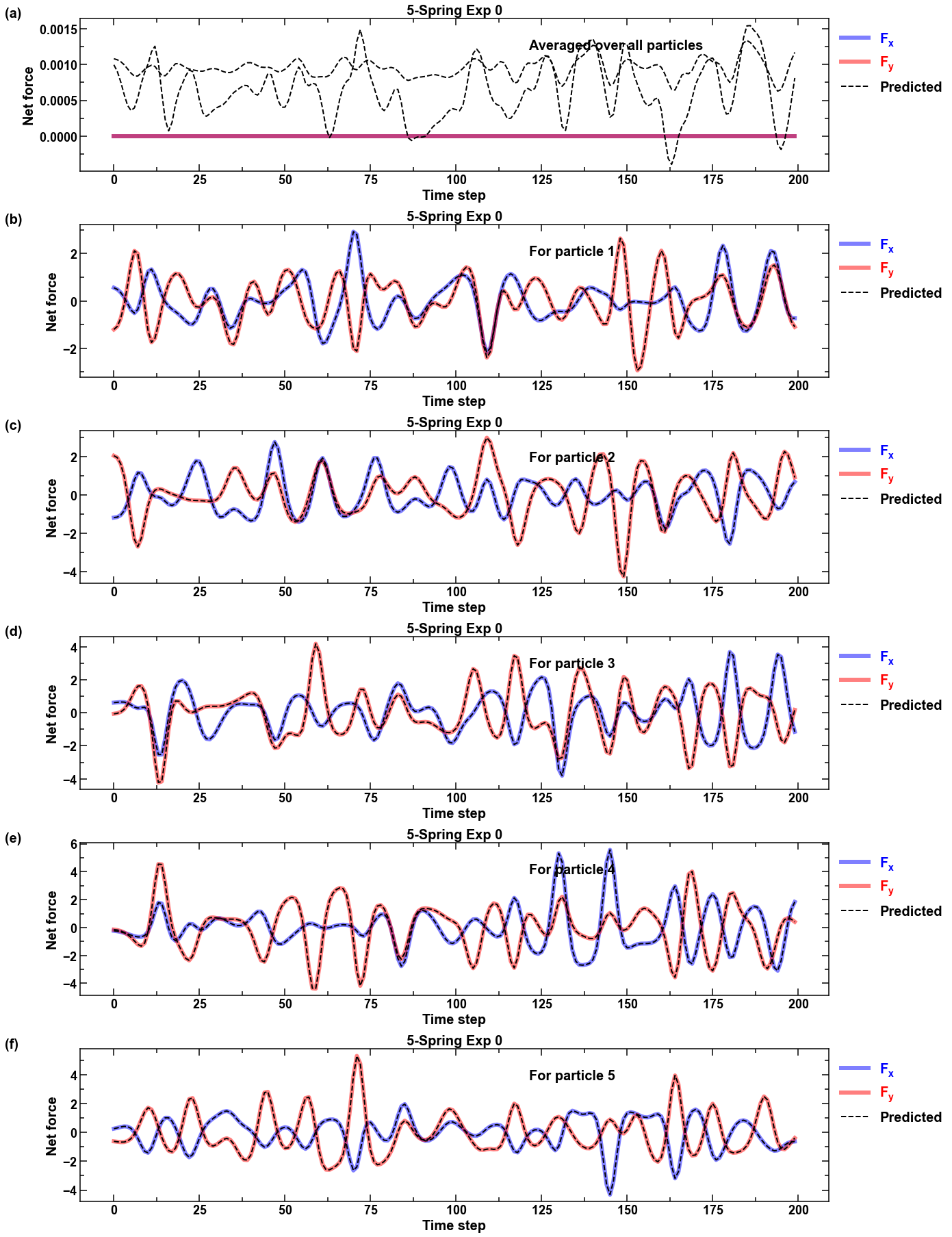}
\caption{\rev{Error in force calculation using \hgn. The solid lines represent the true force and the dashed lines represent the predicted ones. A close overlap among the solid and dashed lines indicate low error. $F_x$ and $F_y$ denote the total forces in the $x$ and $y$ direction respectively. $F_z$ is always 0.}}
\label{fig:fig_b}
\end{figure}

\rev{In this section, we report the errors in the predicted internal and external forces on each particle of a system. Figs.~\ref{fig:fig_c}-\ref{fig:fig_b} presents the results on \gnode, \mcgnode, \lgn and \hgn. As visible, the predicted force (represented using dashed lines) overlaps closely with the actual forces (solid line) in \mcgnode (Fig.~\ref{fig:fig_d}). While the error is also low in \gnode (Fig.~\ref{fig:fig_c}), it is inferior to \mcgnode showcasing the usefulness of the inductive biases. We also note that the error is the highest in \lgn (Fig.~\ref{fig:fig_a}).}

\section{Comparison of \mcgnode, \lgn, \hgn, SymODEN}
\label{app:symoden}
\begin{figure}
\centering
\begin{subfigure}{0.49\columnwidth}
    \includegraphics[width=\columnwidth]{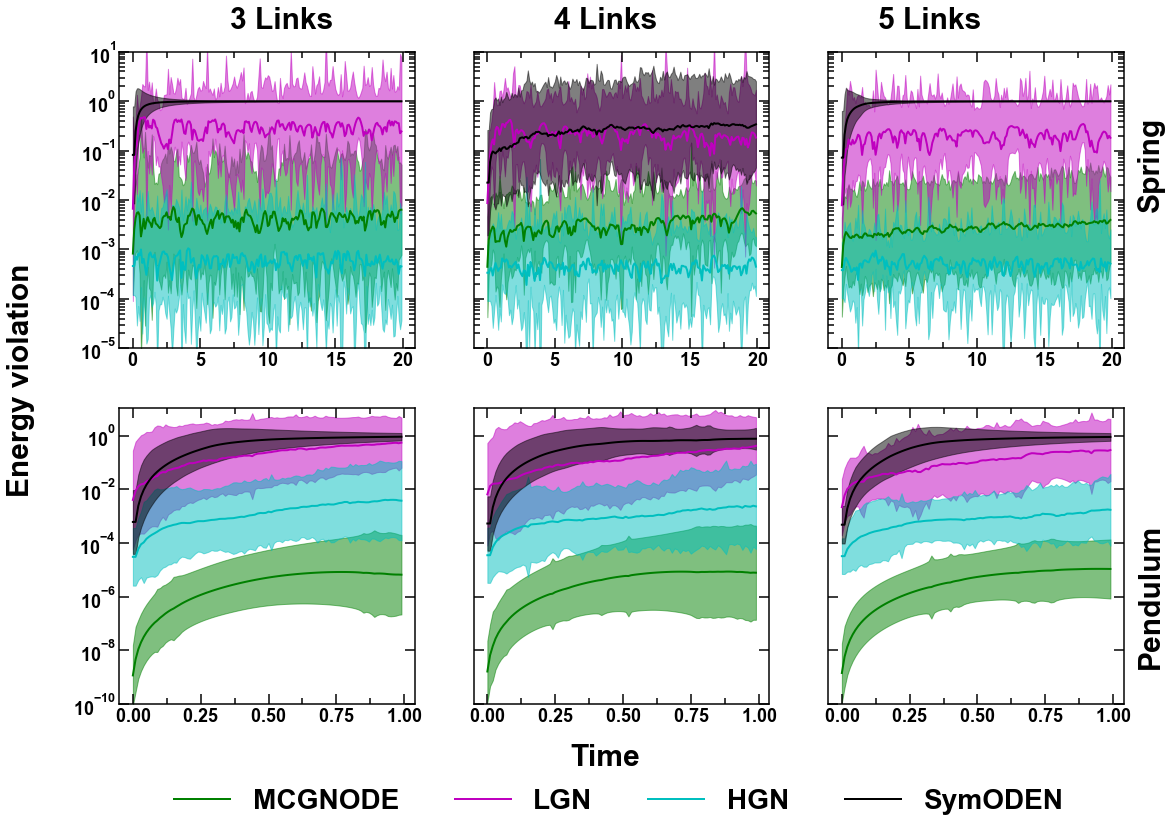}
\end{subfigure}
\begin{subfigure}{0.49\columnwidth}
    \includegraphics[width=\columnwidth]{fig16_Herr.png}
\end{subfigure}
\vspace{-0.10in}
\caption{\rev{Comparison of \mcgnode, \lgn, \hgn, and SymODEN. Note that SymODEN is trained and tested for 3-,4-,5-links system, whereas in case of \mcgnode, \lgn and \hgn training is done only on 5-links system and tested on 3-,4-,5-links. 
}}
\label{fig:fig_16}
\vspace{-0.20in}
\end{figure}

\rev{Fig.~\ref{fig:fig_16} shows the performance of an additional baseline, namely, SymODE-Net (SymODEN)~\cite{zhong2019symplectic} on 3,4,5-spring and pendulum systems. SymODE-Net being a feed-forward MLP is transductive in nature and thus cannot do zero-shot knowledge transfer. Hence, it is trained and tested on each system separately. We observe that the performance of SymODE-Net is comparable to \lgn but poorer than \hgn and \mcgnode.}

\section{Experimental systems}
\label{app:exp_sys}
\subsection{Simulation Environment}
 All the simulations and training were carried out in the JAX environment~\cite{schoenholz2020jax,bradbury2020jax}. The graph architecture was developed using the jraph package~\cite{jraph2020github}. The experiments were conducted on a machine with Apple M1 chip having 8GB RAM and running MacOS Monterey. 

\subsection{$n$-Pendulum}
In an $n$-pendulum system, $n$-point masses, representing the bobs, are connected by rigid bars which are not deformable. These bars, thus, impose a distance constraint between two point masses as 
\begin{equation}
||q_{i}-q_{i-1}|| = l_i   
\end{equation}
where, $l_i$ represents the length of the bar connecting the $(i-1)^{th}$ and $i^{th}$ mass. This constraint can be differentiated to write in the form of a Pfaffian constraint as
\begin{equation}
    q_i\dot{q}_i+q_{i-1}\dot{q}_{i-1}=0
\end{equation}
Note that such constraint can be obtained for each of the $n$ masses considered to obtain the $A(q)$.

The mass of each of the particles is $m_i$ and the force on each bob due to the acceleration due to gravity in $m_ig$. This force will be compensated by the constraint forces, which in turn governs the motion of the pendulum. For \lgn, the Lagrangian of this system has contributions from potential energy due to gravity and kinetic energy. Thus, the Lagrangian can be written as
\begin{equation}
    \mathcal{L}=\sum_{i=1}^n \left(1/2m_i\dot{q_i}^\texttt{T}\dot{q_i}-m_igy_i\right)
\end{equation}
And, for \hgn, the Hamiltonian is given by
\begin{equation}
    \mathcal{H}=\sum_{i=1}^n \left(1/2m_i\dot{q_i}^\texttt{T}\dot{q_i}+m_igy_i\right)
\end{equation}

where $g$ represents the acceleration due to gravity in the $y$ direction.
\subsection{$n$-spring system}
In this system, $n$-point masses are connected by elastic springs that deform linearly with extension or compression. Note that similar to a pendulum setup, each mass $m_i$ is connected only to two masses $m_{i-1}$ and $m_{i+1}$ through springs so that all the masses form a closed connection. Thus, the force experienced by each bob $i$ due to a spring connecting $i$ to its neighbor $j$ is $f_{ij}=1/2k(||q_{i}-q_{j}||-r_{i})$. For \lgn, the Lagrangian of this system is given by
\begin{equation}
    \mathcal{L}=\sum_{i=1}^n 1/2m_i\dot{q_i}^\texttt{T}\dot{q_i}- \sum_{i=1}^n 1/2k(||q_{i-1}-q_{i}||-r_0)^2
\end{equation}
And,for \hgn, the Hamiltonian by:
\begin{equation}
    \mathcal{H}=\sum_{i=1}^n 1/2m_i\dot{q_i}^\texttt{T}\dot{q_i}+ \sum_{i=1}^n 1/2k(||q_{i-1}-q_{i}||-r_0)^2
\end{equation}
where $r_0$ and $k$ represent the undeformed length and the stiffness, respectively, of the spring.

\section{Implementation details}
\label{app:imple}
\subsection{Dataset generation}
\textbf{Software packages:} numpy-1.22.1, jax-0.3.0, jax-md-0.1.20, jaxlib-0.3.0, jraph-0.0.2.dev \\
\textbf{Hardware:}
Chip: Apple M1,
Total Number of Cores: 8 (4 performance and 4 efficiency), 
Memory: 8 GB, 
System Firmware Version: 7459.101.3, 
OS Loader Version: 7459.101.3

For NODE, all versions of \gnode, LGN: All the datasets are generated using the known Lagrangian of the pendulum and spring systems, along with the constraints, as described in Section~\ref{app:exp_sys}. For each system, we create the training data by performing forward simulations with 100 random initial conditions. For the pendulum system, a timestep of $10^{-5}s$ is used to integrate the equations of motion, while for the spring system, a timestep of $10^{-3}s$ is used. The velocity-Verlet algorithm is used to integrate the equations of motion due to its ability to conserve the energy in long trajectory integration. 

While for HGN, datasets are generated using Hamiltonian mechanics. Runge-Kutta integrator is used to integrate the equations of motion due to the first order nature of the Hamiltonian equations in contrast to the second order nature of \lgn, \node, and all versions of \gnode.

From the 100 simulations for pendulum and spring system, obtained from the rollout starting from 100 random initial conditions, 100 data points are extracted per simulation, resulting in a total of 10000 data points. Data points were collected every 1000 and 100 timesteps for the pendulum and spring systems, respectively. Thus, each training trajectory of the spring and pendulum systems are $10s$ and $1s$ long, respectively. Here, we do not train from the trajectory. Rather, we randomly sample different states from the training set to predict the acceleration.\\

\subsection{Architecture}
For \node, we use a fully connected feed-forward neural network with 2 hidden layers, each having 16 hidden units with a square-plus activation function. For all versions of \gnode, graph architecture is used with each MLP consisting of two layers with 5 hidden units. One round of message passing is performed for both pendulum and spring. For \lgn and \hgn, full graph architecture is used with each MLPs consist of two layers with 16 hidden units and one round of message passing were performed. The hyper-parameters were optimized based on the good practices of the model development.

\subsection{Edge features: position vs velocity}
\begin{figure}
\centering
\includegraphics[width=\columnwidth]{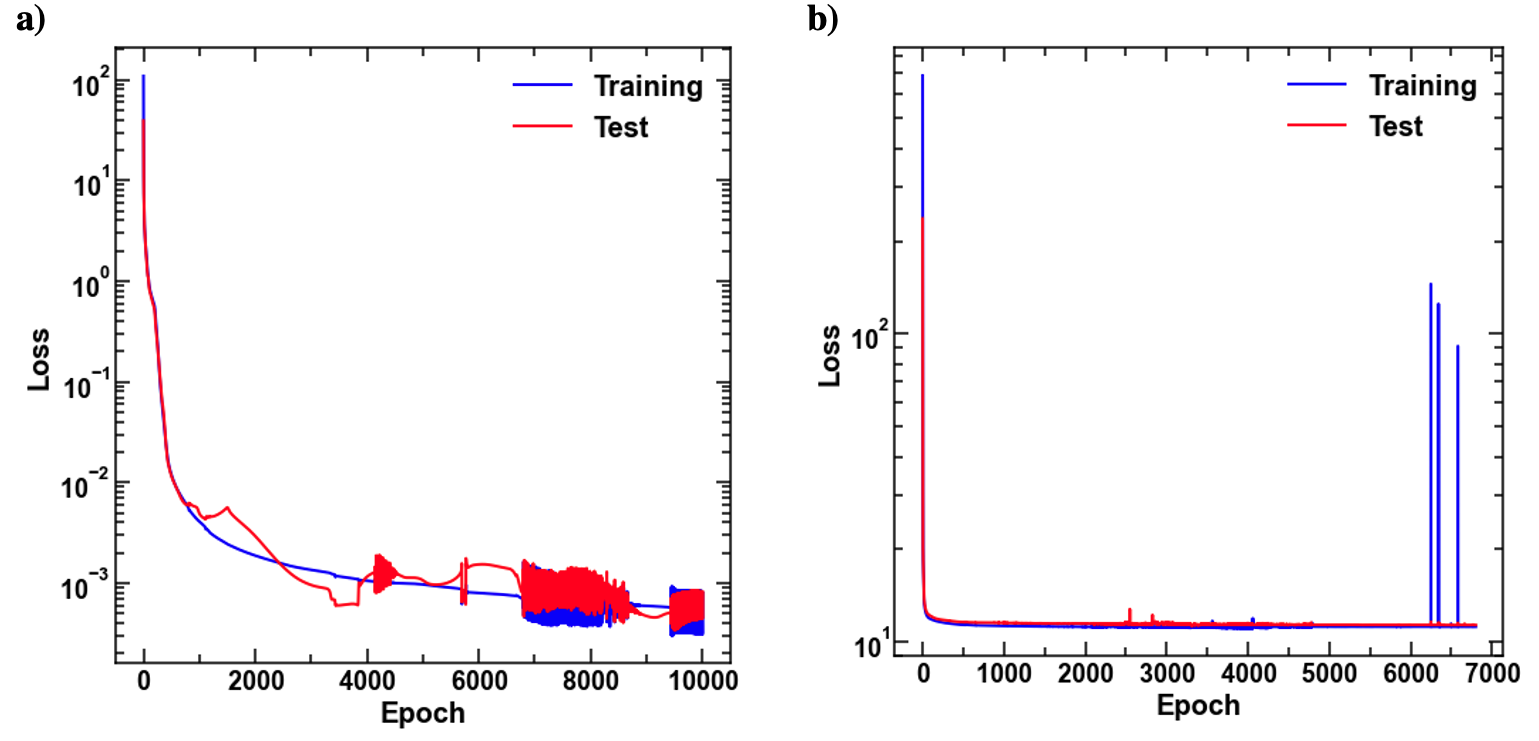}
\caption{Loss curve when (a) relative position is used as edge feature and (b) relative velocity is used as edge feature.}
\label{fig:ablation_study_pos_vel}
\end{figure}

Figure~\ref{fig:ablation_study_pos_vel} shows the loss curve with relative position and relative velocity as edge features. We observe that when relative velocity is provided as the edge feature, model struggles to learn and saturates quickly.

\begin{figure}
\centering
\includegraphics[width=\columnwidth]{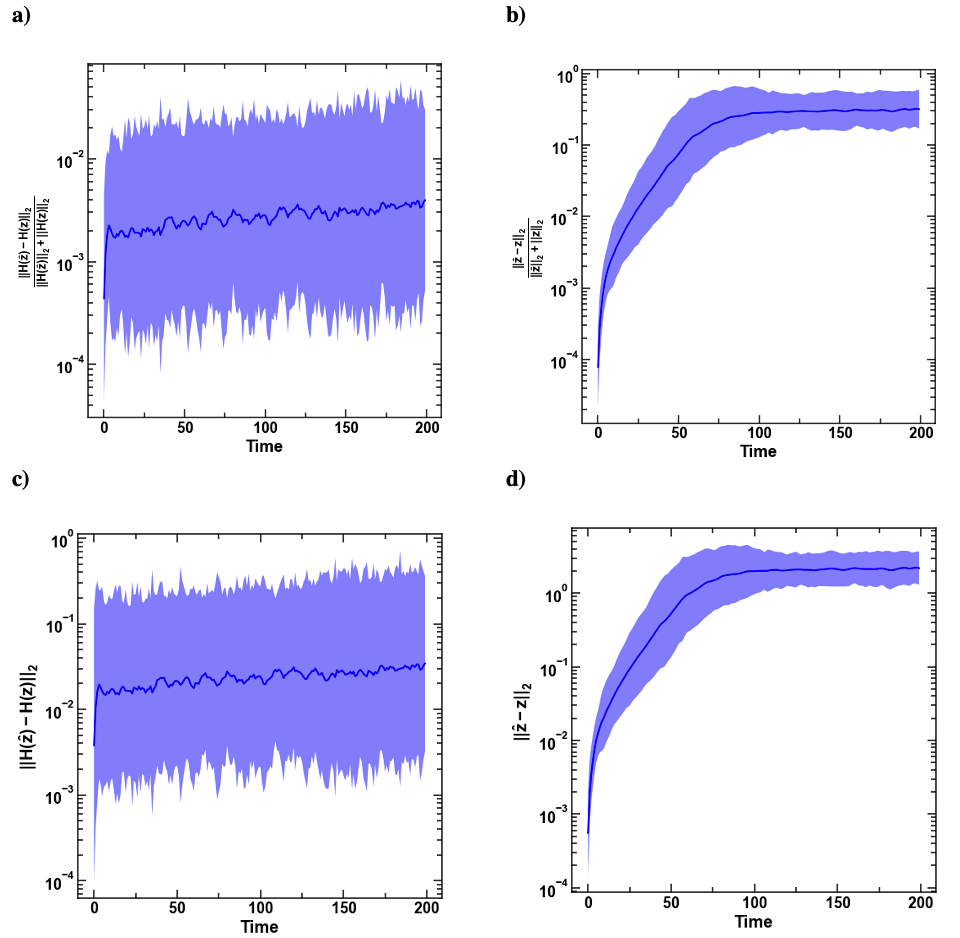}
\caption{Plot comparing (a) relative energy error, (b) relative trajectory error, (c) absolute energy error, (d) absolute trajectory error for the 5-Spring-\mcgnode system}
\label{fig:abs_error}
\end{figure}


\begin{figure}
\centering
\includegraphics[width=\columnwidth]{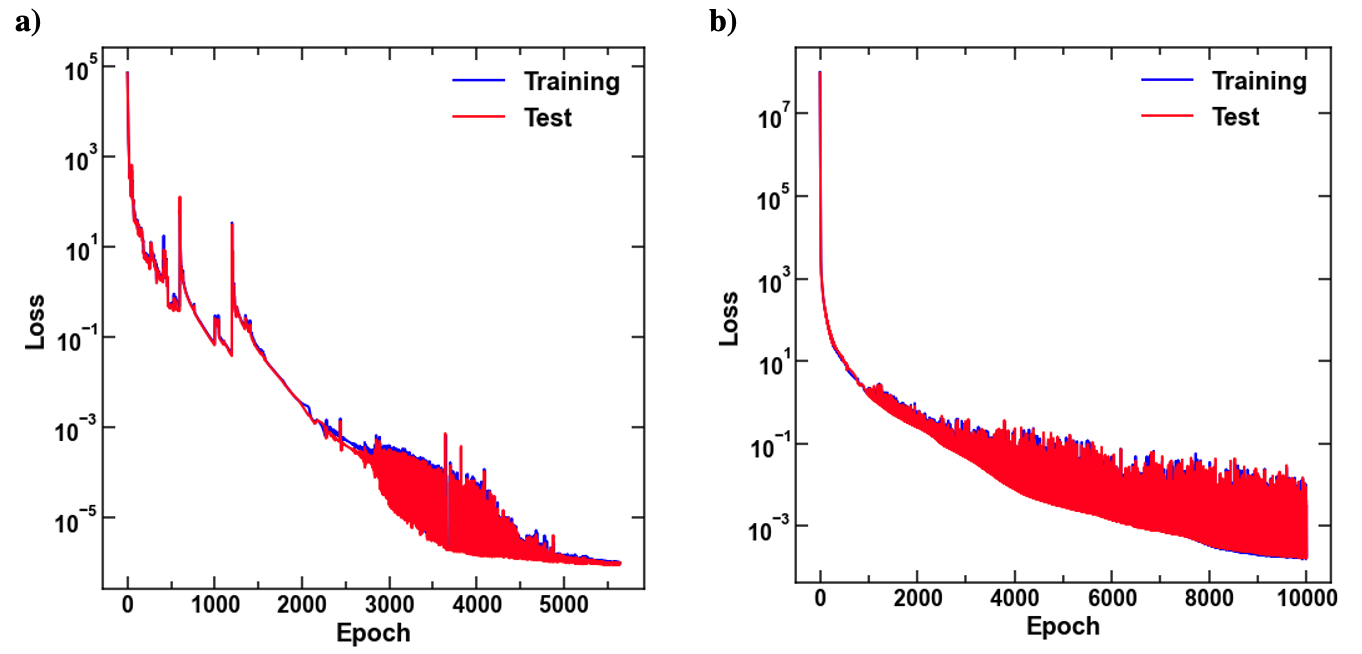}
\caption{Loss curve for peridynamics system using (a) \mcgnode architecture and (b) \gnode architecture.}
\label{fig:peri_gnode_vs_mcgnode}
\end{figure}

\begin{figure}
\centering
\includegraphics[width=\columnwidth]{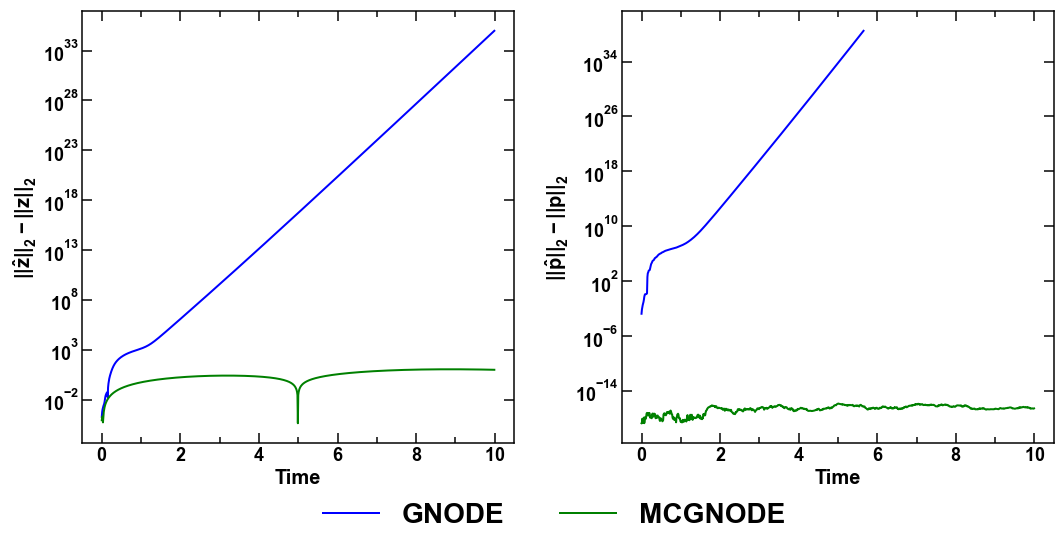}
\caption{Absolute trajectory error error and absolute momentum error  for peridynamics simulation using of \gnode, and \mcgnode.}
\label{fig:abs_peri_gnode_vs_mcgnode}
\end{figure}

\begin{figure}
\centering
\includegraphics[width=\columnwidth]{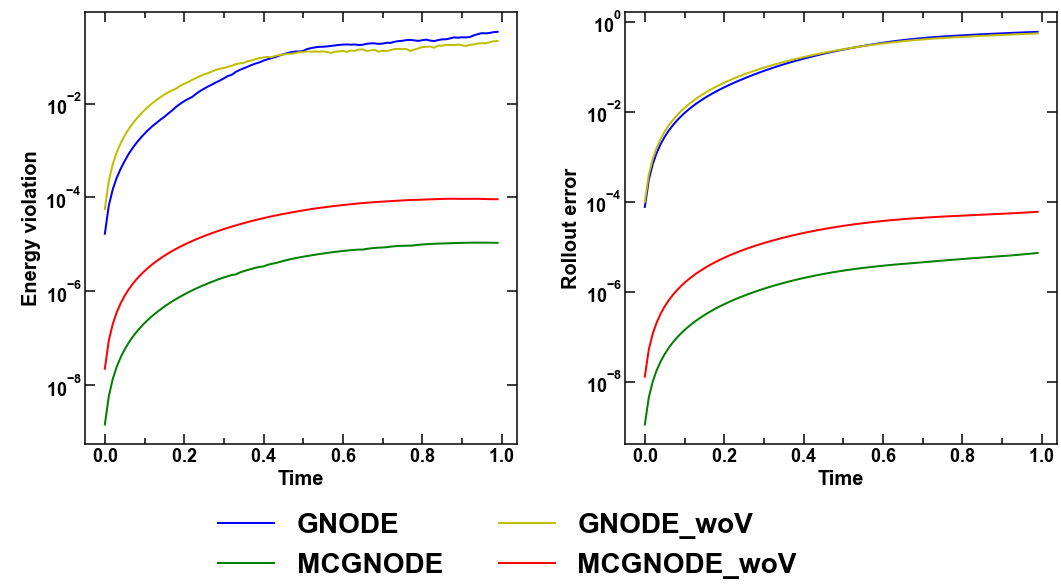}
\caption{Comparison of Energy violation and rollout error for 5-link pendulum using \gnode, \mcgnode, \gnode-woV (\gnode without velocity as input) and \mcgnode-woV (\mcgnode without velocity as input).}
\label{fig:gnode_vs_mcgnode_woV}
\end{figure}

\subsection{Training details}
The training dataset is divided in 75:25 ratio randomly, where the 75\% is used for training and 25\% is used as the validation set. Further, the trained models are tested on its ability to predict the correct trajectory, a task it was not trained on. Specifically, the pendulum systems are tested for $1s$, that is $10^5$ timesteps, and spring systems for $20s$, that is $2\times10^4$ timesteps on 100 different trajectories created from random initial conditions. All models are trained for 10000 epochs. A learning rate of $10^{-3}$ was used with the Adam optimizer for the training.\\

$\bullet$\textbf{NODE}
\begin{center}
\begin{tabular}{ |c|c| } 
 \hline
 \textbf{Parameter} & \textbf{Value} \\ 
 \hline
 Hidden layer neurons (MLP) & 16 \\ 
 Number of hidden layers (MLP) & 2 \\ 
 Activation function & squareplus\\
 Optimizer & ADAM \\
 Learning rate & $1.0e^{-3}$ \\
 Batch size & 100 \\
 \hline
\end{tabular}
\end{center}

$\bullet$\textbf{\gnode, CGNODE, CDGNODE, MC\gnode}
\begin{center}
\begin{tabular}{ |c|c| } 
 \hline
 \textbf{Parameter} & \textbf{Value} \\ 
 \hline
 Node embedding dimension & 5 \\ 
 Edge embedding dimension & 5 \\ 
 Hidden layer neurons (MLP) & 5 \\ 
 Number of hidden layers (MLP) & 2 \\ 
 Activation function & squareplus\\
 Number of layers of message passing & 1\\
 Optimizer & ADAM \\
 Learning rate & $1.0e^{-3}$ \\
 Batch size & 100 \\
 \hline
\end{tabular}
\end{center}

$\bullet$\textbf{LGN, HGN}
\begin{center}
\begin{tabular}{ |c|c| } 
 \hline
 \textbf{Parameter} & \textbf{Value} \\
 \hline
 Node embedding dimension & 8 \\ 
 Edge embedding dimension & 8 \\ 
 Hidden layer neurons (MLP) & 16 \\ 
 Number of hidden layers (MLP) & 2 \\ 
 Activation function & squareplus\\
 Number of layers of message passing & 1\\
 Optimizer & ADAM \\
 Learning rate & $1.0e^{-3}$ \\
 Batch size & 100 \\
 \hline
\end{tabular}
\end{center}

\section{Graph Architectures: \gnode{}, \cgnode{}, \cdgnode{}, \mcgnode{}}
\label{app:diff_graph_arch}
The detailed differences between each of the architectures presented in the work are as follows.\\

$\bullet$ \textbf{GNODE}\\
\textbf{Summary}: Graph version of \node{} that allows inductivity to unseen system sizes.\\
\textbf{Node inputs}: position ($q$), velocity ($\dot{q}$), and type ($t$) \\
\textbf{Edge inputs}: $w_{ij}=(q_i-q_j)$\\
\textbf{Node output}: acceleration of each particle ($\ddot{q}_i$). Note that we do not use Eq.~\ref{eq:gnode_const} here to predict $\ddot{q}$. Instead, we directly predict the acceleration using $\ddot{q}_i=\hat{F}(q,\dot{q},t)$, where $\hat{F}$ is the approximate dynamics learned by the \gnode. Here, predicting force and acceleration are equivalent problems since the acceleration is a simple scaling of the force when the masses of the particles are equal as in the present case.\\
\textbf{Edge output}: Nil.\\
\textbf{Architecture}: Fig.~\ref{fig:graph_architecture}. Here, \gnode{} directly predicts the acceleration and not the force although they are equivalent problems.\\
\textbf{Loss function}: $\mathcal{L}= \frac{1}{n}\left(\sum_{i=1}^n \sum_{t=2}^\mathcal{T} \left(\ddot{q}_i^{\mathbb{T},t}-\hat{\ddot{q}}_i^{\mathbb{T},t}\right)^2\right)$\\
\textbf{Assumed knowns}: None, since we do not use Eq.~\ref{eq:gnode_const} here.\\
\textbf{Inductive bias}: Modeling the physical system as a graph with the particles as nodes and connections between them as edges.\\
 
 $\bullet$\textbf{\cgnode{}}\\
\textbf{Summary}: Same as \gnode{} except for the property that the neural network predicts the force from which acceleration is computed using Eq.~\ref{eq:gnode_const}.\\
\textbf{Node inputs}: Same as in \gnode{}\\
\textbf{Node output}: The force $N_i$ on each particle $i$\\
\textbf{Edge output}: Nil, same as in \gnode{}\\
\textbf{Architecture}: Fig.~\ref{fig:graph_architecture}\\
\textbf{Loss function}: Same as in \gnode. The acceleration is obtained using Eq.~\ref{eq:gnode_const}, where $N$ is obtained from the \cgnode{}, $A(q)$ is assumed to be known, $M$ is the learnable diagonal mass matrix, $\Pi$ is the known external force, $C$ is 0, $\Upsilon$ is the learnable external drag. Note that unless specified, $\Pi$ and $\Upsilon$ is considered to be 0, that is, the system is not subjected to external force or drag. Details below. \\
\textbf{Assumed knowns}: (i) Governing equation for acceleration as given by Eq.~\ref{eq:gnode_const}. (ii) $k$ constraints (holonomic or Pfaffian) in the physical system represented by $A(q) \in \mathbb{R}^{k×D}$ based on the topology and nature (rigid/deformable)) of the system. For instance, the bar in the pendulum is rigid whereas the springs are deformable.\\
\textbf{Inductive bias}: Same as \gnode{} + explicit constraints as given by $A(q)$ and acceleration as given by Eq.~\ref{eq:gnode_const}.\\

$\bullet$\textbf{\cdgnode{}}\\
\textbf{Summary}: Same as \cgnode{} except that the architecture is modified to decouple the computations of internal and body forces. The raw node and edge inputs are the same. However, since we learn two representations per node, global and local, these inputs are used in the following manner:\\
\textbf{Input for global node representations}: position ($q$) and velocity ($\dot{q}$); these features are not involved in message passing. \\
\textbf{Input for local node representations}: type ($t$). \\
\textbf{Input for local edge representations}: $w_{ij}=(q_i-q_j)$, i.e., same as in \gnode. Note that node local input and edge inputs are involved in message passing.\\
\textbf{Node output}: Same as in \cgnode{}, but the local node output and the global node output are concatenated and passed through an MLP to obtain the total node force ($N_i$) as shown in Fig. 1. The intuition is that the local node output represents the internal forces (for instance, the spring forces) and the global node output represents the body forces (for instance, the gravitational force), which when combined gives the total force on each particle. This design thus enables more expressive power for the neural network.\\
\textbf{Edge outputs}: Nil, same as in \cgnode{}.\\
\textbf{Architecture}: Fig. 1.\\
\textbf{Loss function}: Same as in \cgnode{}.\\
\textbf{Assumed knowns}: Same as in \cgnode{}.\\
\textbf{Inductive bias}: Same as in \cgnode{} + decoupling the computation of internal forces (which depend on the topology and is a function of the $w_{ij}$) and body forces (which typically depends on the position of the particle rather than the topology of the system, for instance, gravitational or electromagnetic force).\\

$\bullet$\textbf{\mcgnode{}}\\
\textbf{Summary}: Same as \cdgnode{} except that the architecture is modifed to enforce Newton's third law by incorporation of an additional MLP to predict the individual interaction forces at the edge level.\\
\textbf{Node global input, node local input, edge input}: Same as in \cdgnode{}.\\
\textbf{Edge output}: Force $f_{ij}$ on particle $i$ due to the interaction from the neighboring particle $j$. \\
\textbf{Node output}: Total force $N_i$ for each particle given by $N_i = N_{gi} + \sum_{\mathcal{N}_i}f_{ij}$, where $\mathcal{N}_i$ represents of the set of all the $j$ neighbors of $i$ and $N_{gi}$ represents the force due to external fields obtained by passing the global node output through an MLP.\\
\textbf{Architecture}: Fig.~\ref{fig:graph_architecture} with a difference that there are two MLPs, one to compute $N_{gi}$ which takes as input the global node representation and a second that takes the edge representation as input to predict the $f_{ij}$. We will include a separate figure to explcitly differentiate between \cdgnode{} and \mcgnode{}.\\
\textbf{Loss function}: Same as in \cgnode{}.\\
\textbf{Assumed knowns}: same as in \cgnode{}.\\
\textbf{Inductive bias}: Same as in \cdgnode{} + Newton's third law, that is, the internal forces between the particles are equal and opposite.\\

\begin{figure}
\centering

\includegraphics[width=\columnwidth]{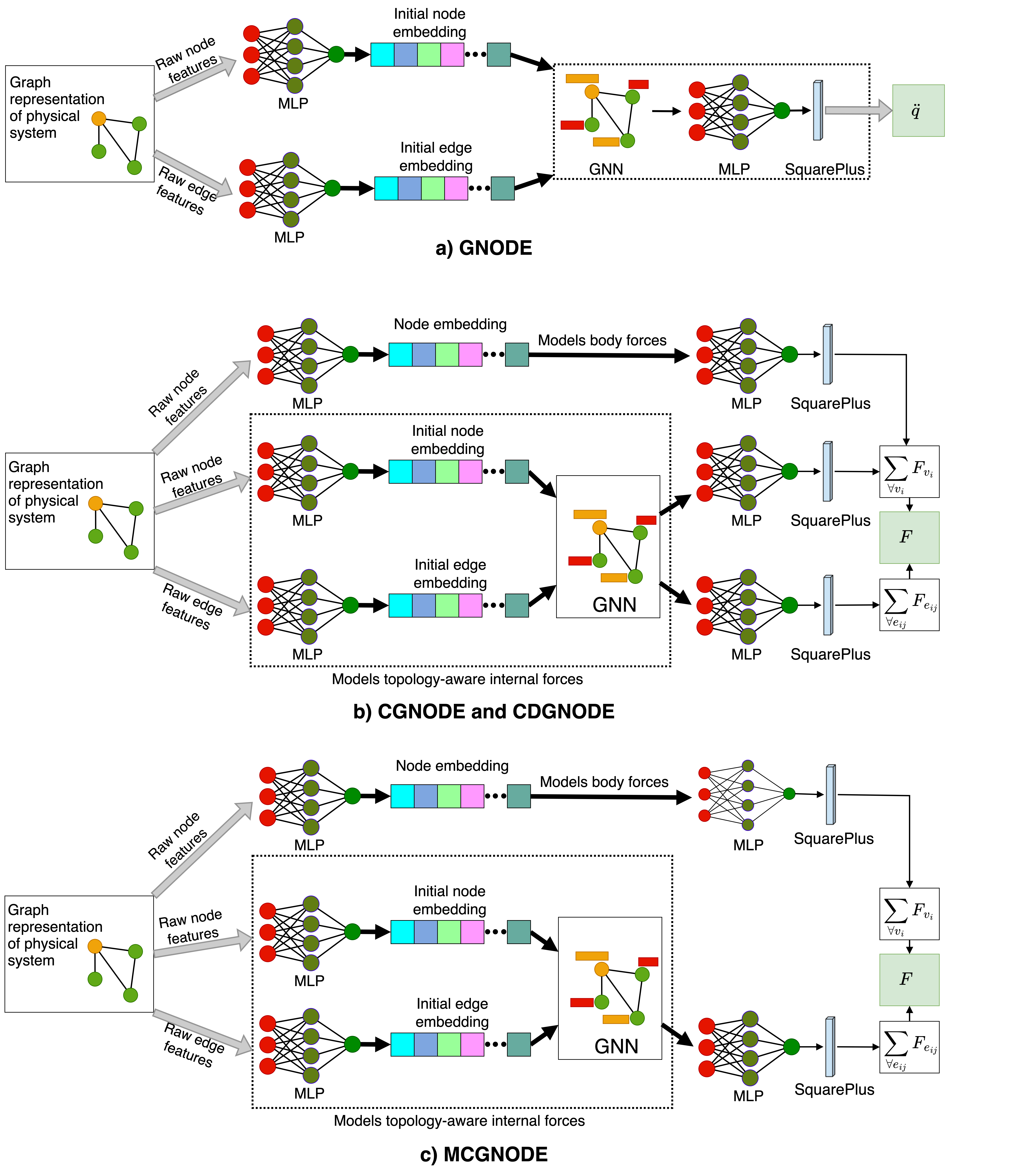}
\caption{Architecture of (a) \gnode, (b) \cgnode{}, \cdgnode{} and (c) \mcgnode{}. Note that the architecture of \cgnode{} and \cdgnode{} are equivalent.}
\label{fig:all_graph_architecture}
\end{figure}

\section{\mcgnode{} vs GNS}
\label{app:gns}

\begin{table}[h]
\centering
\centering
\begin{tabular}{cccc}
\toprule
Time(s) & GNS & GNODE & MCGNODE \\
\midrule
0.00 & 0.74 & 0.92 & 0.26 \\
1.00 & 0.88 & 1.00 & 0.28 \\
2.00 & 0.90 & 1.00 & 0.30 \\
3.00 & 0.89 & 1.00 & 0.36 \\
4.00 & 0.81 & 1.00 & 0.52 \\
5.00 & 0.84 & 1.00 & 0.80 \\
6.00 & 0.88 & 1.00 & 0.81 \\
8.00 & 0.88 & 1.00 & 0.76 \\
9.00 & 0.85 & 1.00 & 0.75 \\
9.90 & 0.84 & 1.00 & 0.70 \\
\bottomrule
\end{tabular}
\caption{Momentum error}
\label{tab:tableME_gns_mcgnode}
\end{table}

\begin{table}
\centering
\begin{tabular}{cccc}
\toprule
Time(s) & GNS & GNODE & MCGNODE \\
\midrule
0.00 & 0.00 & 0.00 & 0.00 \\
1.00 & 1.00 & 0.96 & 0.01 \\
2.00 & 1.00 & 1.00 & 0.02 \\
3.00 & 1.00 & 1.00 & 0.04 \\
4.00 & 1.00 & 1.00 & 0.06 \\
5.00 & 1.00 & 1.00 & 0.06 \\
6.00 & 1.00 & 1.00 & 0.09 \\
7.00 & 1.00 & 1.00 & 0.12 \\
8.00 & 1.00 & 1.00 & 0.16 \\
9.90 & 1.00 & 1.00 & 0.21 \\
\bottomrule
\end{tabular}
\caption{Rollout error}
\label{tab:tableRE_gns_mcgnode}
\end{table}
The tables \ref{tab:tableME_gns_mcgnode}, \ref{tab:tableRE_gns_mcgnode} demonstrate that GNS performs inferior to MCGNode.

\end{document}